\documentclass{article}

\usepackage[main, final]{neurips_2026}

\usepackage[utf8]{inputenc} 
\usepackage[T1]{fontenc}    
\usepackage{hyperref}       
\usepackage{url}            
\usepackage{booktabs}       
\usepackage{threeparttable}
\usepackage{amsfonts}       
\usepackage{nicefrac}       
\usepackage{microtype}      
\usepackage{xcolor}         
\usepackage{graphicx}   
\usepackage{amsmath}
\usepackage{amssymb}
\usepackage{float}
\usepackage{subcaption}
\usepackage{wrapfig}
\usepackage{longtable}
\usepackage{tikz}
\usetikzlibrary{positioning,calc,arrows.meta}
\definecolor{ours}{RGB}{214,96,77}
\definecolor{grp}{RGB}{90,110,140}
\newcommand{\full}{\tikz[baseline=-0.6ex]\fill[black!80] (0,0) circle (2.6pt);}
\newcommand{\half}{\tikz[baseline=-0.6ex]{\draw[black!80,line width=0.6pt] (0,0) circle (2.6pt);
                   \fill[black!80] (0,0) -- (90:2.6pt) arc (90:270:2.6pt) -- cycle;}}
\newcommand{\none}{\tikz[baseline=-0.6ex]\draw[black!80,line width=0.6pt] (0,0) circle (2.6pt);}
\newcommand{\na}{\textcolor{black!35}{--}}

\title{Flow Matching with In-Context Priors for Out-of-Distribution Brain Dynamics}

\author{%
  Sam Gijsen\thanks{\texttt{samgijsen@gmail.com}  \textbullet{}  Code and models: \url{https://github.com/SamGijsen/pinc-flows}} \hspace{1cm}
  Micha\l{} \L{}ukomski \hspace{1cm}
  Marc-André Schulz \hspace{1cm}
  Kerstin Ritter \\
  \\
  Hertie Institute for AI in Brain Health, University of T\"ubingen, Germany \\
  T\"ubingen AI Center, University of T\"ubingen,  Germany \\
  Charit\'e -- Universit\"atsmedizin Berlin, Department of Psychiatry and Psychotherapy, Germany \\
  German Center for Mental Health (DZPG), partner site T\"ubingen, Germany
}

\begin{document}

\maketitle

\begin{abstract}

Flow matching and diffusion models enable conditional generation across domains ranging from images to proteins, with recent extensions to out-of-distribution contexts. Yet generative models of neural time series have largely remained restricted to categorical conditioning, precluding compositional and zero-shot generalization. In this work, we propose PINC-Flow, a per-timestep conditioned diffusion transformer for generating realistic fMRI brain dynamics during unseen cognitive tasks by injecting both compositional language and optional spatial priors in-context. Such zero-shot generation could enable counterfactual neuroscience by supporting in-silico design and evaluation of novel cognitive experiments before empirical validation. Leveraging this model, we evaluate across hundreds of held-out task conditions and characterize predictive performance in relation to the training manifold. From language alone, the model recovers region-specific recruitment across tasks and held-out spatial activation patterns. Spatial priors, when available, complement the text pathway by anchoring generation in regions of task space where language alone degrades, while retaining the compositional structure needed for counterfactual task specification. To our knowledge this is the first generative model of whole-cortex fMRI dynamics for unseen cognitive tasks, advancing counterfactual neuroscience and data-driven experimental design.

\end{abstract}

\section{Introduction}

Generative models trained to map a noise distribution to a target data distribution produce realistic samples across many modalities \citep{sohl2015deep, lipman2022flow}, including images \citep{rombach2021high}, audio \citep{liu2023audioldm}, molecules \citep{hoogeboom2022equivariant}, and proteins \citep{watson2023novo}. Much of the progress in conditional generation has been driven by conditioning on high-dimensional, continuous signals.
Language model embeddings have proven especially effective: they provide a structured semantic space that supports compositional generalization \citep{radford2021learning, saharia2022photorealistic, liu2023audioldm}, while also enabling novel conditioning signals to be specified at inference time. For biological time series, however, it remains unclear how such conditioning can support extrapolation to experimental conditions that were never observed during training.

Such generalization is particularly relevant for biological data: acquisition is costly and candidate interventions vastly outnumber measured examples. The promise is clearly evident in molecular and protein design, where generative models produce realistic samples even away from the direct training support \citep{lee2023exploring, watson2023novo}.
Yet, generative models of neural time series have largely remained restricted to discrete categorical conditioning such as sex, disease labels, or experimental task identity, which prevents the flexible conditioning that enabled generalization in other domains.

We focus on functional MRI, which non-invasively measures a proxy of neural activity \citep{logothetis2001neurophysiological}. The dominant paradigm is task-based: specifically, participants perform structured cognitive experiments (e.g., viewing emotional faces, recalling memories, responding via button presses) while their brain dynamics are recorded over time \citep{barch2013function}. Each experiment is a temporally structured composition of stimuli, instructions, and motor responses. Modeling such experiments as discrete categories conflates this compositional structure into prespecified and fixed labels, foreclosing in-silico investigation of novel cognitive tasks and counterfactual interventions. Task-based fMRI is also acutely affected by data scarcity, as each new study requires expensive scanner time \citep{szucs2020sample}.

No existing approaches address the generation of realistic whole-cortex fMRI dynamics for out-of-distribution cognitive tasks that were never observed during training. In particular, encoding models yield deterministic point estimates of group-mean responses to stimuli, but do not provide temporal dynamics, individual variability, and no native mechanism for novel-task extrapolation. Meanwhile, generative models with categorical conditioning produce realistic dynamics but cannot express experimental structure beyond their fixed label set \citep{vetter2024generating, seo2025scalable, hu2025synthesizing}. What is missing is a generative model that produces realistic neural dynamics and extrapolates compositionally across experiments.

Here we present such a model: a per-timestep conditioned diffusion transformer trained with a flow matching objective under two complementary conditioning regimes (PINC-Flow; Priors-IN-Context). First, we inject compositional language embeddings of each timestep's stimulus, instruction, and response, enabling local and compositional generalization across experimental conditions. We characterize zero-shot generalization across task-space, finding strong text-only generation, albeit breaking down predictably with distance from the training support. Second, we simultaneously train the model to use in-context spatial priors via a self-supervised strategy.  We show that spatial priors anchor generation in regions of task space where language alone degrades, enabling realistic data generation for any experimental design where activation patterns can be specified, while retaining the compositional structure needed for counterfactual task specification. Together, these pathways provide a first step toward generative modeling of whole-cortex fMRI dynamics for unseen cognitive tasks.

\section{Related Work}
  \textbf{Generative models of brain data.}
  Prior work has generated functional connectivity matrices, task-based activation maps, or other aggregate neuroimaging targets rather than full neural time series
  \citep{tavor2016task,gal2022act,serin2025generating}. Recent diffusion models generate brain dynamics more directly \citep{zhai2025brain,hu2025synthesizing,seo2025scalable,xia2026brain, tew2025t2idiff,gao2026braincast}, but typically
  condition on fixed metadata such as demographics, diagnosis, or task identity, if at all. These models can synthesize realistic samples within observed labels, but cannot specify unseen tasks compositionally at
  inference time.

  \textbf{Continuous conditioning in neuroimaging.}
  Encoding models predict brain responses from stimulus features, spanning hand-engineered methods \citep{naselaris2011encoding} to pretrained language and vision representations
  (TRIBEv2, \cite{d2026foundation}; Appendix \ref{app:extra_related_work}). Pretrained representations are used by decoding models to infer or reconstruct observed stimuli
  \citep{ozcelik2023natural,scotti2023reconstructing,bosch2025brain} or in training foundation models to aid generalization \citep{gijsen2025eeg,wei2025fmri}. While these works show that continuous representation spaces can support generalization, they target deterministic prediction, stimulus reconstruction, or representation learning rather than diverse generation of task-fMRI dynamics under novel experimental designs.

  \textbf{Structured conditioning for biological extrapolation.}
  Structured conditioning has enabled useful extrapolation in biological domains including protein design \citep{watson2023novo,ingraham2023illuminating}, molecular generation
  \citep{igashov2024equivariant,schneuing2024structure}, and single-cell perturbation modeling \citep{lotfollahi2019scgen,lotfollahi2023predicting}. These settings share the central pressure of task fMRI:
  experiment spaces are combinatorial, acquisition is expensive, and directly observed conditions sparsely cover the space of scientific interest.

\section{Methods}

\begin{figure}[t]
    \centering
    \begin{subfigure}{0.73\textwidth}
        \centering
        \includegraphics[width=\linewidth]{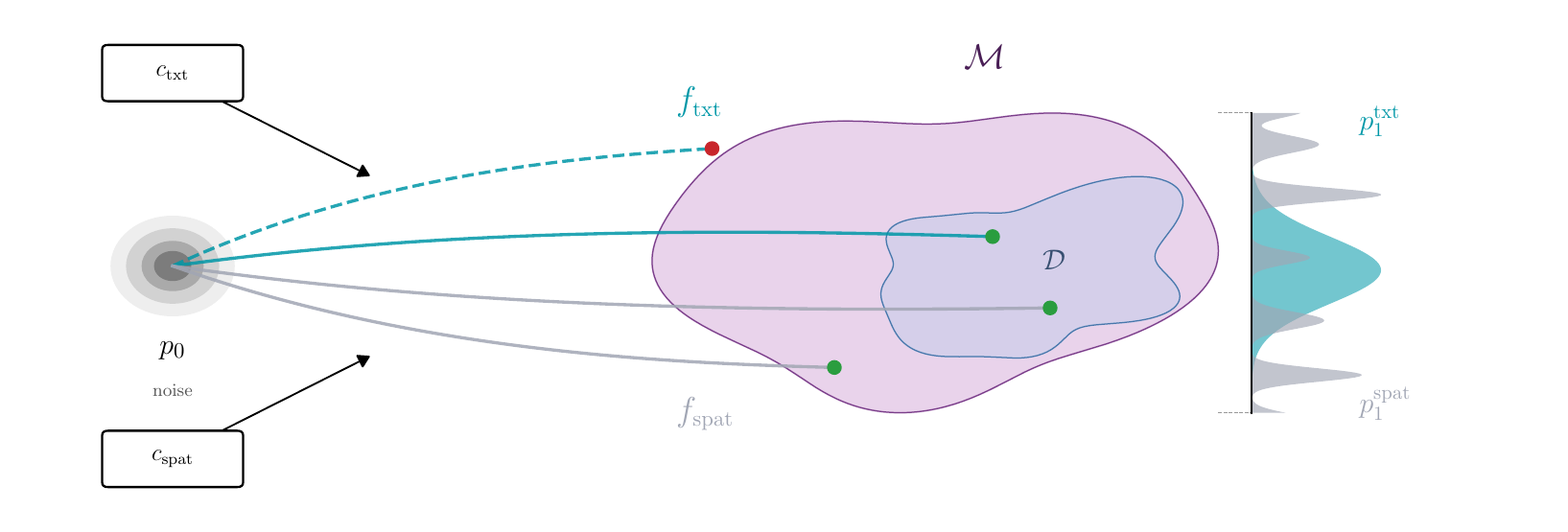}
        \caption{}
        \label{fig:manifold}
    \end{subfigure}
    \hfill
    \begin{subfigure}{0.25\textwidth}
        \centering
        \includegraphics[width=\linewidth]{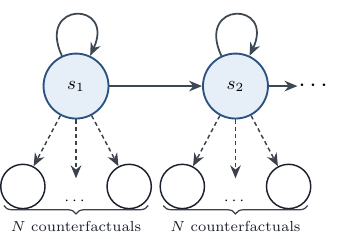}
        \caption{}
        \label{fig:markov}
    \end{subfigure}
    \caption{a) The flow model is trained to enable two conditional sampling functions $f_{\text{txt}}$ and $f_{\text{spat}}$, with distinct induced densities. Solid paths denote intended transports to valid trajectories on $\mathcal{M}$; the dashed path illustrates a possible failure mode under unsupported text conditioning. b) To alleviate overfitting to sequence ordering during training, we occasionally present counterfactual sequences.}
\end{figure}

\begin{figure}[!t]
    \centering
    \includegraphics[width=0.90\linewidth]{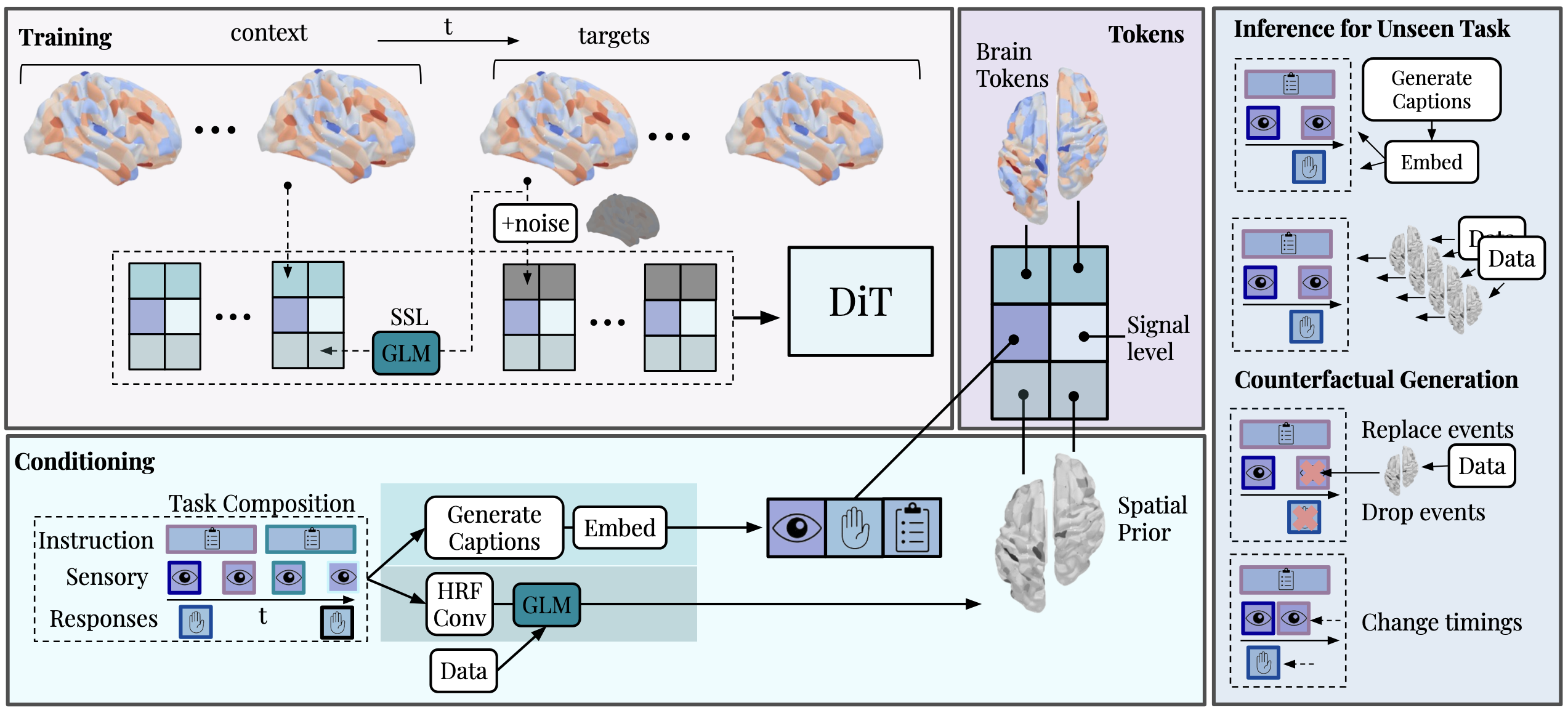}
    \caption{We train a diffusion transformer (DiT) to sample trajectories of brain dynamics. The model is conditioned on context volumes, text embeddings of task captions, and spatial priors. Both conditioning pathways enable compositional generalization by decomposing experimental tasks.}
    \label{fig:methods_model_conditioning}
\end{figure}

We aim to model the distribution of brain dynamics that a participant would exhibit during any plausible cognitive experiment. Let \(\mathcal{M}\) denote the manifold of valid task-fMRI trajectories induced by the space of such experiments, and let \(\mathcal{D}_{\mathrm{train}} \subset \mathcal{M}\) denote the finite collection of trajectories captured by our training data. Although \(\mathcal{D}_{\mathrm{train}}\) covers only a small portion of \(\mathcal{M}\), two structural properties make generalization beyond it tractable. First, experimental conditions are compositional: each is a structured combination of stimuli, instructions, and motor responses, components that recur across experiments and enable natural language descriptions. Second, both the temporal and spatial signature of activation for a given cognitive operation have a significant shared structure between studies and subjects. We exploit both properties through a dual conditioning scheme depicted in Figure~\ref{fig:manifold}. A text-guided sampler \(f_{\text{txt}}\) uses the compositional structure of natural-language descriptions, supporting local and compositional extrapolation across \(\mathcal{M}\). However, when queried with unsupported or far-extrapolative text conditions, \(f_{\text{txt}}\) may fail to extrapolate within \(\mathcal{M}\), instead producing trajectories that leave the manifold of valid task-fMRI dynamics. A spatially guided sampler \(f_{\text{spat}}\) instead conditions on an explicit activation map, helping to anchor generation within a constrained region of \(\mathcal{M}\) while requiring the activation pattern to be provided or estimated externally. Together, the two pathways trade compositional flexibility against distributional anchoring and provide complementary coverage of \(\mathcal{M}\).

We instantiate the above as a generative model of fMRI dynamics with three design choices. First, whereas prior work conditions diffusion models through global mechanisms such as adaptive layer-norm, which do not allow per-timestep resolution, we instead insert conditioning tokens directly into the transformer context. This allows the cognitive content of each timestep to be specified independently and enables counterfactual interventions on event timing and composition. Second, both conditioning pathways operate in continuous, structured spaces, making local and compositional generalization possible at inference. Third, we train the model end-to-end with a flow matching objective, yielding realistic and diverse trajectories rather than deterministic point predictions. We detail the architecture and training procedure below.

\subsection{Generative Model}

We model parcellated fMRI sequences. At each timestep $t$, $\mathbf{x}_t \in \mathbb{R}^{P}$ represents the activation of all $P$ parcels, with $x_{t,p}$ denoting parcel $p$. We split $\mathbf{x}_t$ into two hemispheric vectors $\mathbf{x}^h_t \in \mathbb{R}^{P/2}$ for $h \in \{L, R\}$. A training sample spans $T = T_c + T_g$ timesteps, with $T_c$ real-data context volumes and $T_g$ generation volumes. The latter are noised during training and iteratively denoised at inference. Hemisphere-specific encoders $E_L, E_R$ project $\mathbf{x}^h_t \mapsto \mathbf{z}^h_t \in \mathbb{R}^D$, whereas the decoders $\mathrm{Dec}_L, \mathrm{Dec}_R$ map back to parcel space.

\textbf{Architecture.}
We use a Diffusion Transformer \citep{peebles2023scalable} over the flat token sequence of length $6T$ consisting of six tokens per timestep: 
\[
\bigl\{(\mathbf{z}^L_t,\, \mathbf{z}^R_t,\, \mathbf{c}^{\text{spat},L}_t,\, \mathbf{c}^{\text{spat},R}_t,\, \mathbf{c}^{\text{txt}}_t,\, \mathbf{c}^{\sigma}_t)\bigr\}_{t=1}^{T}\] to which we add a learnable temporal embedding $\mathbf{e}_t$ and token-type embedding. The \emph{brain tokens} $\mathbf{z}^h_t$ are the hemisphere-encoded activations for $t \leq T_c$, and the noised target for $t > T_c$. The \emph{spatial-prior tokens} $\mathbf{c}^{\text{spat},h}_t = E_h(\boldsymbol{\beta}^h_t)$ project a per-timestep spatial prior map $\boldsymbol{\beta}^h_t \in \mathbb{R}^{P/2}$ (defined below) through the same hemispheric encoder used for brain tokens; this shared projection ensures that brain-state and spatial-prior tokens occupy a common representational space. 

The \emph{language token} $\mathbf{c}^{\text{txt}}_t$ is constructed compositionally from three captions describing the cognitive content at $t$: a sensory caption (e.g., ``visual: photos of human faces''), an instruction caption (``choose the most attractive option''), and a response caption (``right thumb button press''). Each caption is embedded with a pretrained sentence transformer \citep{bge_embedding}, linearly projected to dimensions $0.4D$, $0.4D$, $0.2D$, and concatenated to form a single language token in $\mathbb{R}^D$. By embedding the three components independently, the language token decomposes the experimental description along its natural axes of variation, allowing recombinations unseen during training. We randomly mask components during training. To improve robustness to specific phrasings, we generate six phrase variants per caption with a language model (Appendix \ref{app:llm}) and sample randomly. Finally, the \emph{signal-level token} $\mathbf{c}^{\sigma}_t$ embeds the scalar signal level $\sigma_t$. To enable classifier-free guidance at inference, language tokens, spatial-prior tokens, and context volumes are independently dropped during training.

We condition on natural-language descriptions rather than modality-specific stimulus encoders (e.g., CLIP image embeddings, audio encoders). This unifies the sensory, instruction, and response components of a task in a single embedding space and scales to datasets in which raw stimulus files are unavailable (e.g. the Nakai dataset introduced below).

\textbf{Self-supervised spatial prior.}
The spatial-prior pathway is trained self-supervised, exploiting a fundamental property of the fMRI signal. Neural activity drives a slow hemodynamic response function $hrf(t)$, well-modeled as a canonical double-gamma kernel peaking at ${\approx}\,5$\,s and decaying over ${\approx}\,15$\,s \citep{friston1998event}. The general linear model (GLM) routinely used in fMRI analysis models the signal as
\[
x_{t,p} \;=\; \sum_{k} \beta_{p,k}\,(hrf * s_k)(t) \;+\; \varepsilon_{t,p},
\]
where $s_k(t)$ is the onset indicator of event type $k$ and $\beta_{p,k}$ is a per-parcel regression coefficient. Because the HRF spreads each event's effect over the upcoming temporal window, the event coefficient encodes the spatial pattern expected to modulate the upcoming response window. Concretely, for an event of type $k$ with an onset at timestep $t$, we refer to the coefficients $\{\boldsymbol{\beta}_{p,k}\}$ over parcels $p$ belonging to hemisphere $h$ as the vector $\beta^h_t \in \mathbb{R}^{P/2}$. To train $f_\text{spat}$, we supply $\boldsymbol{\beta}^h_t$ to the model at the matching timestep, and the GLM is fit on the subject's own run. Because the prior is derived from the same data the model reconstructs, no external labels are required, and learning to use the prior reduces to a well-posed self-supervised objective.

We estimate $\boldsymbol{\beta}^h_t$ per event type rather than jointly across events, facilitating counterfactual edits to the event composition at inference. To alleviate overfitting to specific task signatures and to promote sample diversity, we estimate $\boldsymbol{\beta}^h_t$ per subject rather than per task. At inference, the spatial prior can be dropped, predicted from text, or supplied from external empirical sources such as prior task-fMRI studies. In our evaluation, priors estimated from held-out recordings define an oracle empirical setting. This setting is not a zero-resource prediction benchmark like \(f_{\text{txt}}\), but tests the intended use of \(f_{\text{spat}}\). Namely, spatially anchored generation without retraining, where supplied or recombined activation patterns are realized as realistic temporal dynamics under event, response, and timing compositions unseen during training.

\textbf{Flow matching.}
Let $z = \{\mathbf{z}^h_t : h \in \{L,R\}, t > T_c\}$ denote the collection of brain tokens at the generation timesteps, and let $c^{\text{ctx}}=\{\mathbf{z}^h_t\}_{t\leq T_c}$ denote the context volumes. We group the conditioning as $c = \left( c^\text{txt}, c^\text{spat}, c^\text{ctx}\right)$. We learn a velocity field $v_\theta(z_\tau, \tau, c)$ that transports $p_0 = \mathcal{N}(0, I)$ to a conditional data distribution $p_1(\cdot \mid c)$ along the linear probability path $z_\tau = (1-\tau)\,z_0 + \tau\,z_1$ with FM time $\tau \in [0,1]$ \citep{lipman2022flow}.  
The network uses an endpoint-prediction parameterization: it predicts
\(\hat z_1 = g_\theta(z_\tau,\tau,c)\), and the velocity is computed as \(v_\theta = (\hat z_1 - z_\tau)/(1-\tau)\).
Training draws $\tau \sim \mathcal{U}(0,1)$, $z_0 \sim p_0$, and minimizes $\lVert \hat z_1 - z_1 \rVert^2$ on generation timesteps. At inference we generate samples by $K$-step Euler integration, $z_{\tau + \Delta\tau} = z_\tau + \Delta\tau\, v_\theta(z_\tau, \tau, c)$ with $\Delta\tau = 1/K$. Appendix \ref{app:stochastic_fc} describes a stochastic variant used only for calibration analyses and Appendix \ref{app:fm_parameterization} ablates the endpoint parameterization.

\textbf{Conditional samplers.}
The two pathways introduced above correspond to running the learned dynamics
under different conditioning sets, inducing distinct approximate transports
from $p_0$ to pathway-specific conditional endpoint distributions:
\begin{align*}
    f_{\text{txt}}  &: p_0 \to
    p_\theta\!\left(z_1 \mid c^{\text{txt}}, c^{\text{ctx}}\right)
    \approx p_1^{\text{txt}}\!\left(z_1\right)
    := p_1\!\left(z_1 \mid c^{\text{txt}}, c^{\text{ctx}}\right), \\
    f_{\text{spat}} &: p_0 \to
    p_\theta\!\left(z_1 \mid c^{\text{txt}}, c^{\text{spat}}, c^{\text{ctx}}\right)
    \approx p_1^{\text{spat}}\!\left(z_1\right)
    := p_1\!\left(z_1 \mid c^{\text{txt}}, c^{\text{spat}}, c^{\text{ctx}}\right).
\end{align*}
$f_{\text{txt}}$ relies on compositional language conditioning alone;
$f_{\text{spat}}$ additionally anchors generation to an explicit activation map.

\textbf{Classifier-free guidance.}
Independent dropout of language, spatial-prior, and context tokens during training allows $v_\theta$ to be evaluated using arbitrary subsets of $c$ \citep{ho2022classifier}. At inference we group the conditioning into two channels: a \emph{task} channel $c^{\text{task}}$, which is $c^{\text{txt}}$ under $f_{\text{txt}}$ and $(c^{\text{txt}}, c^{\text{spat}})$ under $f_{\text{spat}}$, and the \emph{context} channel $c^{\text{ctx}}$. 
When a channel is omitted, we replace its conditioning tokens with a learned null token, denoted here \(\varnothing\).
With task scale $w_{\text{task}}$ and context scale $w_{\text{ctx}}$, each Euler step combines three forward passes,
\[
\tilde v \;=\; w_{\text{task}}\, v_\theta(c^{\text{task}}, c^{\text{ctx}}) \;+\; (w_{\text{ctx}} - w_{\text{task}})\, v_\theta(\varnothing, c^{\text{ctx}}) \;+\; (1 - w_{\text{ctx}})\, v_\theta(\varnothing, \varnothing),
\]
equivalent to the unconditional velocity plus a context contribution $w_{\text{ctx}}\bigl(v_\theta(\varnothing, c^{\text{ctx}}) - v_\theta(\varnothing, \varnothing)\bigr)$ and a task contribution $w_{\text{task}}\bigl(v_\theta(c^{\text{task}}, c^{\text{ctx}}) - v_\theta(\varnothing, c^{\text{ctx}})\bigr)$.

\textbf{Counterfactual training.}
Task-fMRI runs contain fixed condition orderings, so the model can overfit to observed condition
  transitions. We diversify training by splicing discontinuous segments from the same subject (Figure \ref{fig:markov}),
  creating novel condition-to-condition transitions (for a more detailed treatment, see Appendix \ref{app:counterfactual_interp}). Because splicing introduces an artificial
  signal discontinuity, we downweight the loss immediately after the splice and inform its restoration trajectory based on the hemodynamic response function. We provide the precise weighting scheme in Appendix \ref{app:counterfactual_training}.

\section{Experimental Setup}
Whereas models typically are assessed on a handful of tasks, both training a model capable of generalizing \emph{across} tasks and performing a robust evaluation requires unique datasets. To this end, we focus on including a large set of varied experimental conditions. This enables evaluation along two axes. First, we verify the generated time series themselves to investigate realism in terms of spectral content, variance, autocorrelation, functional connectivity, and correspondence with real data. Second, we investigate the model's capability to generate data conditioned on tasks unseen during training.  

\textbf{Data.}
We combine four complementary datasets spanning large subject cohorts and broad task coverage. HCP provides large-scale task-fMRI across seven tasks \citep{barch2013function}, IBC provides dense within-subject coverage across 53 tasks \citep{ponce2026individual}, Nakai adds 103 rapidly alternating task conditions for training \citep{nakai2020quantitative}, and UK Biobank contributes large-scale resting-state data for unconditional training \citep{miller2016multimodal}. All data are parcellated with the Schaefer-400 atlas, resampled to 1\,Hz, low-pass filtered at 0.25\,Hz, and parcel-wise z-scored.

We evaluate zero-shot task generation with seven task-held-out folds. Each fold holds out one HCP task and approximately eight IBC tasks, such that every task is held out once; the dynamics model is
trained from scratch for each fold. For HCP, we additionally hold out 100 subjects across all tasks, enabling simultaneous evaluation of held-out-task and held-out-subject generalization. Dataset details are provided in Appendix \ref{app:data_details}.

\begin{table}[t]
\footnotesize
  \caption{Out-of-distribution prediction evaluated by comparing GLM $\beta$ weights from the real data with those recovered from synthetic data. $n$ are number of conditions or contrasts. Map: spatial map used for conditioning (Direct: Direct's prediction; Cross-run: map from the other phase-encoding acquisition; Oracle: empirical map). We show mean$\pm$std across conditions/contrasts.}
  \label{main_results}
  \centering
  \setlength{\tabcolsep}{3.5pt}
  \begin{tabular}{@{}llcccccc@{}}
    \toprule
    & & \multicolumn{3}{c}{HCP} & \multicolumn{3}{c}{IBC} \\
    \cmidrule(lr){3-5} \cmidrule(lr){6-8}
    Model & Map & $n$ & Single Cond. & Task Contrasts & $n$ & Single Cond. & Task Contrasts \\
    \midrule
    \multicolumn{8}{@{}l}{\emph{Zero-resource}} \\
    Direct                    & --        & 17/13 & 0.280 $\pm$ 0.269 & 0.262 $\pm$ 0.375 & 191/60 & 0.245 $\pm$ 0.230 & 0.240 $\pm$ 0.243 \\
    kNN retrieval             & kNN       & 17/13 & 0.295 $\pm$ 0.308 & 0.308 $\pm$ 0.286 & 191/60 & 0.301 $\pm$ 0.268 & 0.317 $\pm$ 0.316 \\
    Text-FIR                  & --        & 17/13 & 0.482 $\pm$ 0.312 & 0.249 $\pm$ 0.226 & 191/60 & 0.281 $\pm$ 0.331 & 0.381 $\pm$ 0.373 \\
    PINC-Flow-Unconditioned   & --        & 17/13 & 0.435 $\pm$ 0.407 & 0.039 $\pm$ 0.219 & 191/60 & 0.000 $\pm$ 0.197 & 0.017 $\pm$ 0.161 \\
    PINC-Flow-$f_\text{txt}$  & --        & 17/13 & \textbf{0.601} $\pm$ 0.313 & \textbf{0.346} $\pm$ 0.212 & 191/60 & \textbf{0.418} $\pm$ 0.306 & \textbf{0.469} $\pm$ 0.324 \\
    PINC-Flow-$f_\text{spat}$ & Direct    & 17/13 & 0.534 $\pm$ 0.251 & 0.293 $\pm$ 0.221 & 191/60 & 0.378 $\pm$ 0.247 & 0.407 $\pm$ 0.300 \\
    \midrule
    \multicolumn{8}{@{}l}{\emph{Emperical prior}} \\
    PINC-Flow-$f_\text{spat}$ & Cross-run & 17/13 & 0.846 $\pm$ 0.131 & 0.584 $\pm$ 0.184 & -- & -- & -- \\
    PINC-Flow-$f_\text{spat}$ & Oracle    & 17/13 & 0.883 $\pm$ 0.128 & 0.749 $\pm$ 0.158 & 191/60 & 0.641 $\pm$ 0.211 & 0.766 $\pm$ 0.172 \\
    \midrule
    \multicolumn{8}{@{}l}{\emph{In-distribution}} \\
    PINC-Flow-$f_\text{spat}$ & Oracle    & 17/13 & 0.916 $\pm$ 0.146 & 0.889 $\pm$ 0.094 & 191/60 & 0.700 $\pm$ 0.220 & 0.839 $\pm$ 0.133 \\
    \bottomrule
  \end{tabular}
\end{table}

\textbf{Evaluation.}
For task-level evaluation, we generate synthetic counterparts of held-out recordings and analyze them with the same GLM used for the real data. Because the model generates windows rather than entire runs in one pass, we tile each held-out recording with non-overlapping generation windows of length \(T_g\), conditioning each window on the task conditioning and leakage-controlled context volumes. The resulting synthetic sequence is then analyzed as a run-level recording. The model receives sensory, instruction, and response captions at every timestep. 
The evaluation GLM follows standard fMRI practice and jointly fits event regressors: each event regressor is weighted by its duration or fractional occupancy within a volume and convolved with the canonical HRF. 
We fit identical design matrices to matched real and synthetic samples, yielding contrast maps \(\boldsymbol{\beta}_k\) and \(\hat{\boldsymbol{\beta}}_k\), which we compare across parcels. For brevity, we use ``contrast'' to refer both to single event maps and differences between event maps.

To prevent task information from leaking through the context window, context volumes are offset from the generated task segment. When available, we use pre-task periods with fixation conditioning; otherwise, we sample subject-specific context from a different task and mask its task conditioning. Task annotations are provided only for generated timesteps. Real and synthetic GLMs are always fit on matched samples and matched events (further details are given in Appendix \ref{app:glm_evaluation}). For \(f_\text{txt}\), we provide only text embeddings as task conditioning. For \(f_\text{spat}\), we additionally evaluate three spatial-prior regimes: predicted priors from the Direct model, independent cross-run (estimated on a second run of the held-out task, mimicking a pilot, localizer, or repeated session), and oracle empirical priors (estimated on the evaluated recording), none of which are seen by the dynamics model during training. We refer to this latter setting as \(f_\text{spat}\) unless otherwise specified.

\textbf{Baseline:} To the best of our knowledge, no prior generative model targets out-of-distribution task-fMRI dynamics. We therefore construct  multiple baselines. First, we use a strong text-to-contrast baseline (\emph{Direct}; Appendix \ref{app:direct_baseline}) that predicts contrast maps directly from the same language conditioning used by our model. Direct provides both an evaluation anchor for held-out contrast recovery and, when no empirical spatial prior is available, an alternative source of spatial priors for the dynamics model. Unlike our dynamics model, however, Direct predicts static contrast maps rather than full time series, and therefore cannot model temporal dynamics, sample diversity, individual variability, or counterfactual changes to event timing. 

Next, we add two families of generative models (Appendix \ref{app:generative_baselines}). HRF-VAR convolves an activation map with the held-out event timings and adds a residual process using PCA/VAR(1). We pass the map predicting by \emph{Direct} (Sim-Direct), a nearest-neighbor map (from the training set; Sim-kNN), or the same priors used for PINC-Flow. Text-FIR is a deterministic linear finite-impulse-response encoder which maps from text embeddings to BOLD. As these latter two baselines output time series, we use the identical evaluation protocol as for PINC-Flow. Finally, we also evaluate our model without task conditioning and fully in-distribution as reference points.

\paragraph{Hyperparameters}
We use hemispheric encoders to keep the total token sequence length compact despite the multiple token types injected for each timestep. These, as well as the decoders are learned linear projections. We use the Schaefer-400 atlas \citep{schaefer2018local} similar to recent foundation modeling efforts \citep{dong2024brain,lane2025scaling,gijsen2026brain}. Our default parameters are as follows. We use $T_c=8$, $T_g=16$, which yields 144 total tokens per window. To train PINC-Flow, we train a DiT backbone with 12 layers and $D=1024$ for 100K steps (compute use is described in Appendix \ref{app:compute_usage}). At inference, we perform 64 Euler steps with guidance scales $w_{\text{task}}=w_\text{ctx}=1.5$. We provide further optimization and model parameters in Appendix \ref{app:hyperparameters}.

\section{Results}

\paragraph{Generated trajectories recover activation patterns of unseen task conditions.}
We first ask whether trajectories generated under unseen task conditioning recover the activation structure, by comparing \(\hat{\boldsymbol{\beta}}_k\) for condition $k$ recovered from real and generated trajectories.
Critically, these activation maps are not predicted directly by the model, but rather are recovered from sampled fMRI trajectories. As such, successful contrast recovery requires the model to generate temporally structured task-evoked dynamics rather than merely outputting static activation maps.

\begin{figure}[t]
    \centering
    \includegraphics[width=1\linewidth]{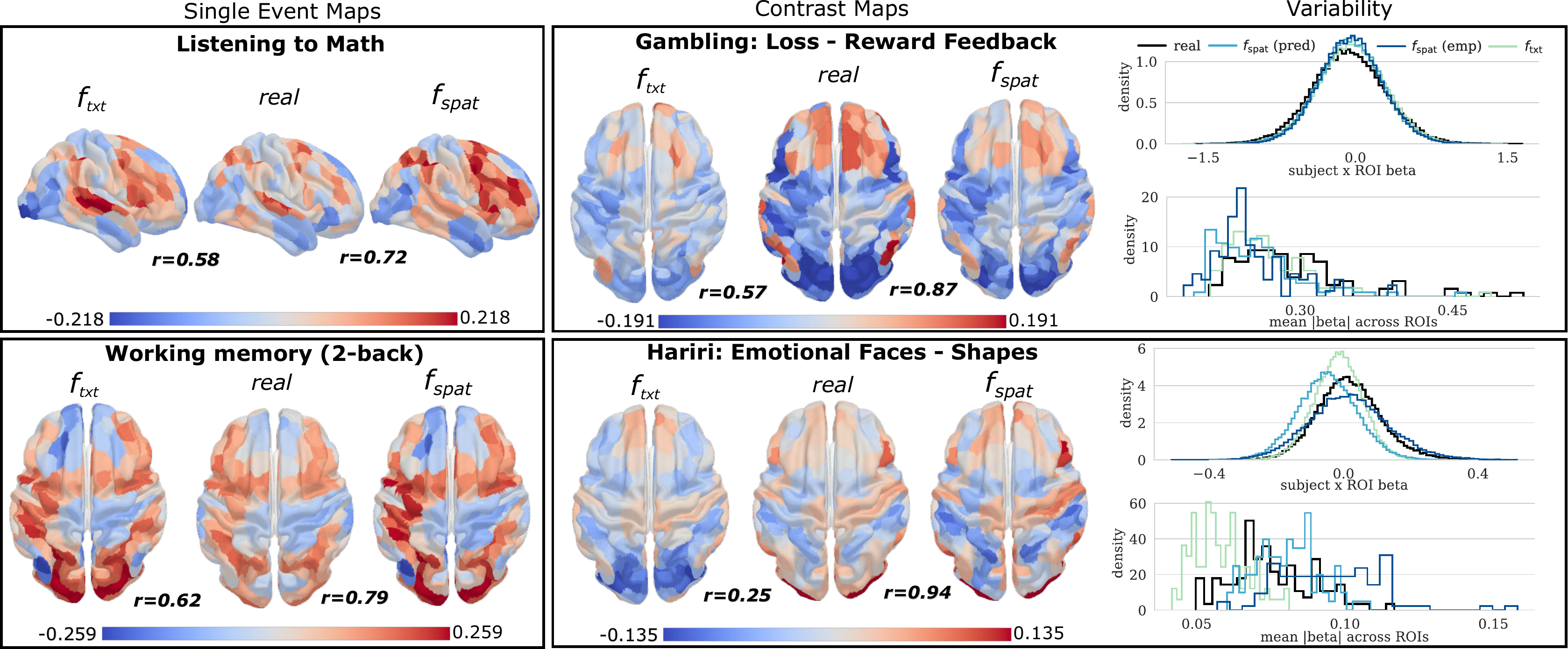}
    \caption{Example $\beta$-maps recovered from generated timeseries under unseen task conditioning. We show median cases as well as a failure case in the bottom right, where the contrast map by $f_{\text{txt}}$ was not able to capture emotional processing during a perceptual masking task.}
    \label{fig:main_vis}
\end{figure}

The resulting OOD results are shown in Table \ref{main_results} with visualizations in Figure \ref{fig:main_vis}. Text-only generation performs strongly under held-out task conditioning: PINC-Flow-\(f_{\text{txt}}\) reaches \(r=0.60\) for HCP single conditions and \(r=0.47\) for IBC task contrasts. It thereby performs better than the baselines in all four settings. Empirical but unseen spatial oracle priors further improve performance, with PINC-Flow-\(f_{\text{spat}}\) reaching \(r=0.88\) and \(r=0.77\), respectively. Cross-run priors are modestly less effective, but remain considerably stronger than $f_\text{txt}$. These empirical prior conditions should be interpreted as a spatial-anchoring test rather than a zero-resource prediction benchmark: the supplied maps provide the target spatial hypothesis, but the model must still realize it as temporally structured fMRI trajectories under held-out event, response, and timing compositions. We evaluate in-distribution as well, to provide an approximate upper bound on performance, as it implicitly accounts for contrast reliability and shared evaluation idiosyncrasies (e.g. generating tiled windows).

We note that the large standard deviations result primarily from a diverse task set, rather than noisy model evaluation. For example, the difference between PINC-Flow-$f_\text{txt}$ and the Direct baseline is significant with $t(280; \text{pooled; paired over tasks}) = 11.70$, $p<10^{-25}$. The unconditioned PINC-Flow model performs well on HCP single conditions due to the model generating prototypical visual task activations, which maps well to some of its conditions with rapid visual presentation.

\begin{table}[t]
\centering
\footnotesize
\begin{minipage}[t]{0.46\textwidth}
  \centering
  \setlength{\tabcolsep}{4pt}
  \caption{Spatial prior vs.\ generated recording (HCP, $n{=}30$ pooled conditions and contrasts). $q$: $r$(prior, target) ; $f$: $r$(simulator output, prior).}
  \label{tab:prior_sim}
  \begin{tabular}{@{}lcccc@{}}
    \toprule
    & & \multicolumn{2}{c}{HRF--VAR sim.} & PINC-Flow-$f_\text{spat}$ \\
    \cmidrule(lr){3-4} \cmidrule(lr){5-5}
    Prior source & $q$ & $f$ & $r$ & $r$ \\
    \midrule
    Direct               & .274 & .995 & .273 & \textbf{.430} \\
    $k$NN retrieval  & .318 & .973 & .287 & --            \\
    \midrule
    Cross-run        & .654 & .991 & .644 & \textbf{.733} \\
    Oracle empirical & .738 & .995 & .728 & \textbf{.825} \\
    \bottomrule
  \end{tabular}
\end{minipage}%
\hfill
\begin{minipage}[t]{0.51\textwidth}
  \centering
  \setlength{\tabcolsep}{3pt}
  \caption{Trajectory similarity for HCP OOD generation. Median [5th, 95th pct.]}
  \label{tab:temporal}
  \begin{tabular}{@{}lcc@{}}
    \toprule
    & Trajectory $r$ & Paired $\Delta$ \\
    \midrule
    Real split-half ref. & .942 [.935, .950] & -- \\
    \midrule
    \multicolumn{3}{@{}l}{\emph{Text-conditioned}} \\
    Text-FIR                & .354 [.344, .364] & \\
    PINC-Flow-$f_\text{txt}$      & \textbf{.500} [.476, .519] & $+$.145 [.123, .162] \\
    \midrule
    \multicolumn{3}{@{}l}{\emph{Oracle prior}} \\
    HRF--VAR sim.           & .571 [.550, .590] & \\
    PINC-Flow-$f_\text{spat}$     & \textbf{.620} [.602, .643] & $+$.050 [.027, .074] \\
    \bottomrule
  \end{tabular}
\end{minipage}
\end{table}

\begin{figure}[t]
    \centering
    \begin{subfigure}{0.27\textwidth}
        \centering
        \includegraphics[width=\linewidth]{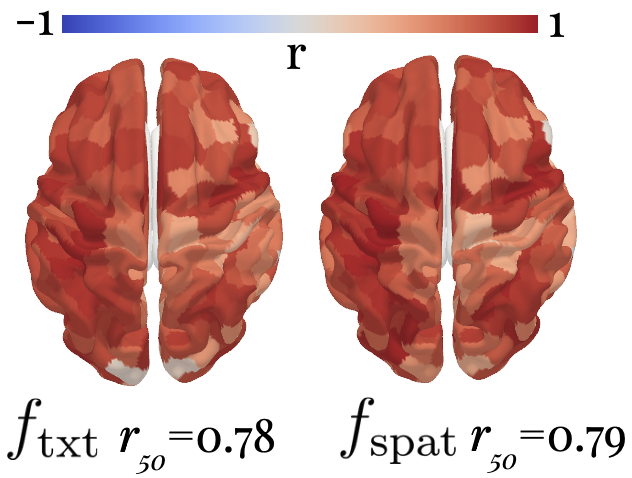}
        \caption{}
        \label{fig:r_across_tasks}
    \end{subfigure}
    \hfill
    \begin{subfigure}{0.71\textwidth}
        \centering
        \includegraphics[width=\linewidth]{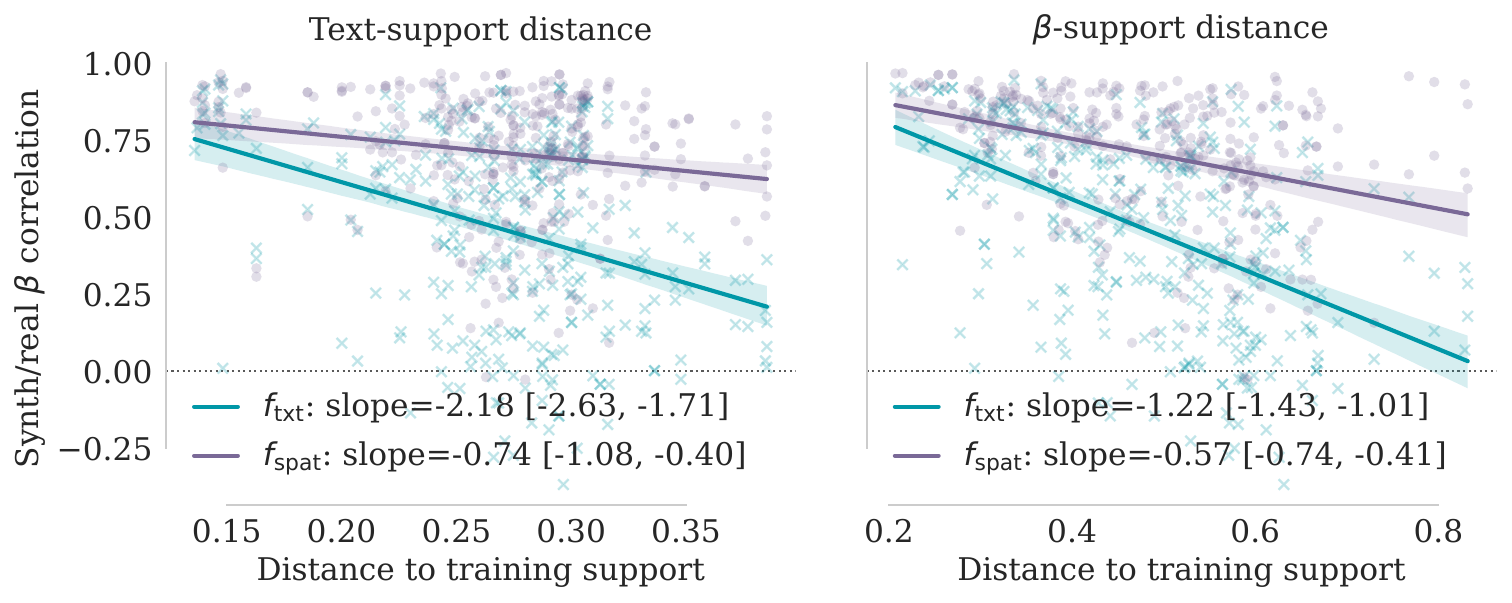}
        \caption{}
        \label{fig:support_distance_slopes}
    \end{subfigure}
    \caption{a) Per-region recruitment across tasks. b) The two conditional samplers are compared in their OOD performance as a function of the distance to the training manifold.}
    \label{fig:r_across_tasks_and_slopes}
\end{figure}

Beyond asking whether each generated task map matches the corresponding real task map, we also investigate whether individual regions are recruited appropriately across unseen tasks. For each region, we correlate predicted and real \(\beta\) values across held-out tasks. This evaluates whether the model recovers region-specific functions over the OOD task space. We observe high correlations for both \(f_{\text{txt}}\) (median \(r=0.78\)) and \(f_{\text{spat}}\) (median \(r=0.79\); Figure~\ref{fig:r_across_tasks}). 

\paragraph{Generated dynamics exceed baselines and their priors.} We find that the HRF-VAR simulator effectively passes through the maps it receives, as the correlation between its predictions and the target is consistently capped by the correlation between the supplied prior and the target (Table \ref{tab:prior_sim}). Meanwhile, PINC-Flow exceeds the prior in each regime (+.16 predicted, +.08 cross-run, +.10 oracle). These results indicate that its advantage is not reducable to retrieval of neighboring tasks or construction of GLM-compatible signals from a supplied prior map.

\paragraph{Temporal trajectories.} We compare the similarity of generated trajectories for held-out HCP tasks at the (Yeo-17) network level (Table \ref{tab:temporal}). PINC-Flow-$f_\text{txt}$ exceeds Text-FIR given identical captions ($\Delta = +.15 [.12, .16]$), and the $f_\text{spat}$ branch exceeds HRF-VAR when both use the oracle prior ($\Delta = +0.05 [.03, .07]$). All models scoring considerably lower than the real split-half reference implies that significant progress is still to be made.

\paragraph{Spatial priors stabilize extrapolation away from the training manifold.}
We next investigate the relationship between the predictability of unseen task conditioning and its distance to the training manifold, pooling single conditions and task contrasts. We compute distance via nearest-neighbor distance to the training manifold in both text and \(\beta\)-space (Figure~\ref{fig:support_distance_slopes}). As expected, OOD performance degrades as targets move further from the training manifold. However, this degradation is considerably muted for \(f_{\text{spat}}\) compared to \(f_{\text{txt}}\). This supports its intended role: when text conditioning becomes unreliable far from the training manifold, explicit spatial hypotheses can anchor generation while preserving the ability to compose event content, responses, and timing at inference.

\paragraph{Generated samples exhibit realistic time-series statistics and subject structure.}
 
\begin{figure}[t]
    \centering
    \includegraphics[width=0.99\linewidth]{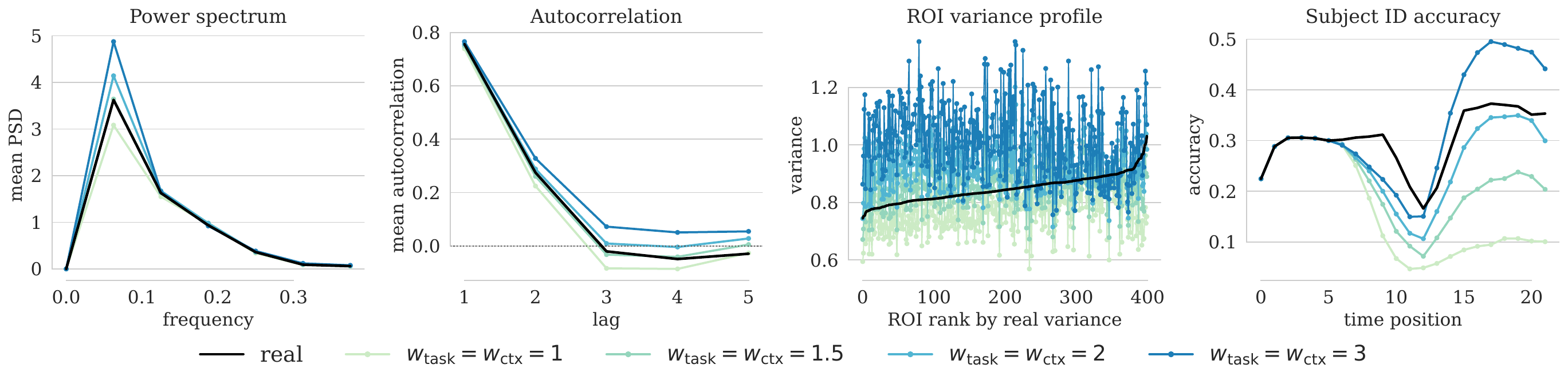}
    \caption{General statistics of the generated data distribution and subject recoverability.}
    \label{fig:basic_stats_with_id}
\end{figure}

Because the primary evaluation above reduces generated trajectories to GLM contrasts, we next verify that the model also captures broader properties of the time-series distribution under unseen conditioning. We observe that our model produces realistic data in terms of power spectrum, autocorrelation, and variance (Figure~\ref{fig:basic_stats_with_id}). Naturally, these statistics depend on the guidance parameters \(w\), which guide generation to emphasize context volumes, task conditioning, or both. We find that the default setting \(w_{\text{task}}=w_{\text{ctx}}=1.5\) produces sensible statistics.

\begin{figure}[t]
    \centering
    \includegraphics[width=0.99\linewidth]{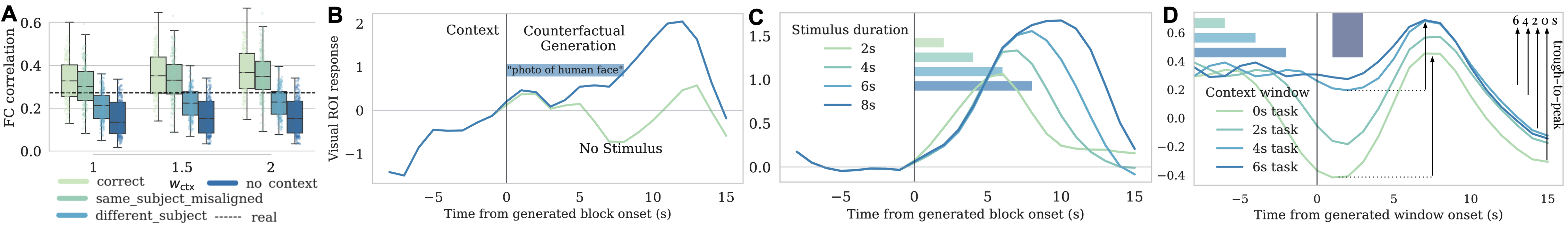}
    \caption{A) Functional connectivity correlations under different context regimes (mean$\pm$std). Counterfactual generations are shown for single- (B) and averaged-trajectories (C). D) Longer task presentation in context induces parametric repetition-inhibition effects.}
    \label{fig:counterfactual}
\end{figure}

We then ask whether context volumes steer generation toward subject-specific dynamics. A subject-identification classifier trained on real held-out-subject HCP windows recovers subject identity from generated samples, and identity evidence increases with context guidance (Figure~\ref{fig:basic_stats_with_id}). Functional connectivity provides a more calibrated test: agreement is highest for the correct context, lower for same-subject misaligned context, lower again for different-subject context, and lowest without context (Figure~\ref{fig:counterfactual}A). This ordering indicates that the model uses context to capture subject-specific structure rather than merely producing generic task dynamics. At high guidance, generated FC can become overly similar to the conditioning context; in Appendix~\ref{app:stochastic_fc}, we show that stochastic latent sampling reduces this over-conditioning while preserving the expected context ordering.

\paragraph{Counterfactual temporal interventions.}
We study whether the model has learned temporally structured task responses rather than only static spatial associations. We use counterfactual generation to test this, varying both future task conditioning and the content of the context window. The model produces HRF-like responses to counterfactual stimulus presentation, and generated trajectories are sensitive to the recent context history. Interestingly, longer task durations in the context lead to reduced subsequent responses, indicating the model learned dynamics consistent with repetition inhibition (Figure~\ref{fig:counterfactual}; \cite{larsson2012fmri}). This supports per-timestep task conditioning as a mechanism for counterfactual interventions on event timing and composition. 

We provide an ablation on our splicing training strategy in Appendix \ref{app:splicing_ablation}. We find that it particularly helps novel transitions involving in-distribution task conditions. Specifically, splicing removes the tendency to regress towards expected (as per the training distribution) rather than requested conditions.

\begin{figure}[t]
    \centering
    \begin{subfigure}{0.64\textwidth}
        \centering
        \includegraphics[width=\linewidth]{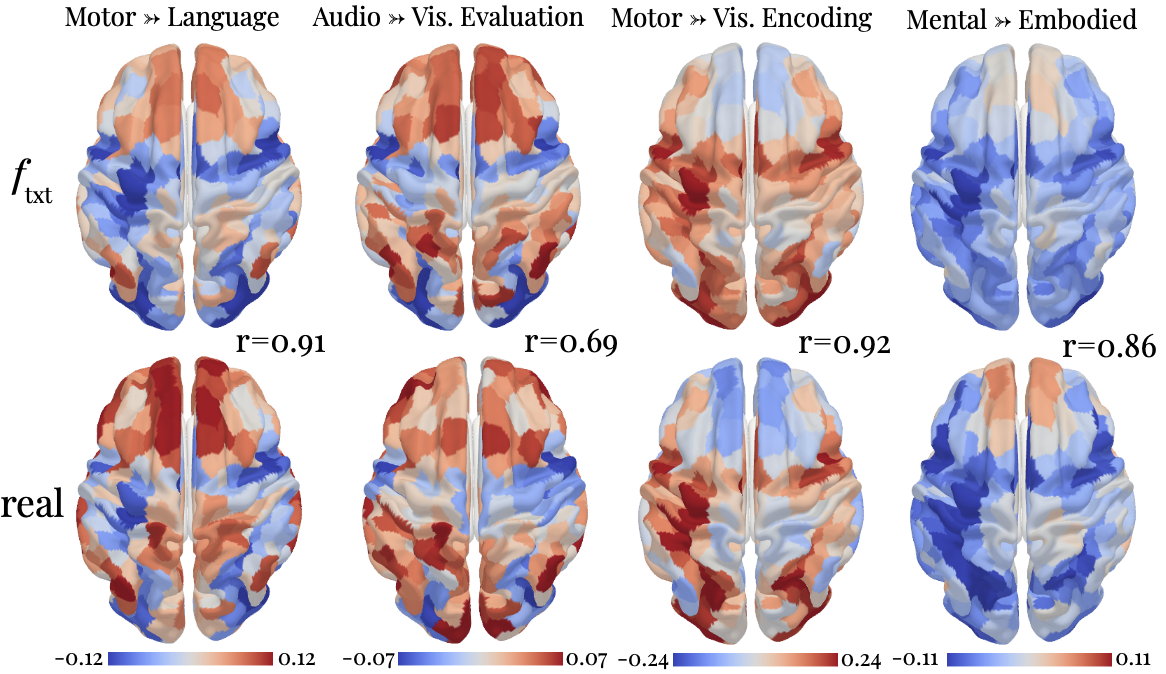}
        \caption{Coefficient maps resulting from regressing $\beta$-values onto task space PCA scores.}
        \label{fig:semantic_tuning}
    \end{subfigure}
    \hfill
    \begin{subfigure}{0.30\textwidth}
        \centering
        \includegraphics[width=\linewidth]{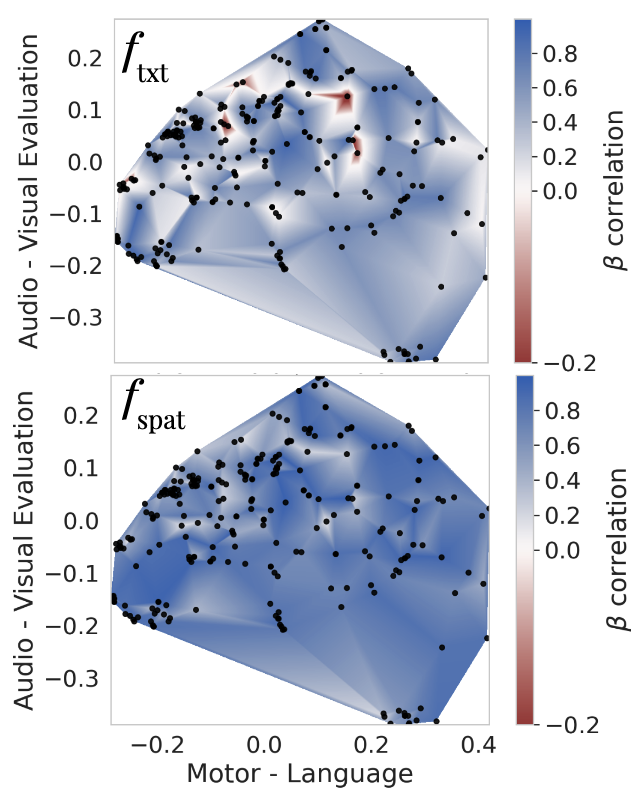}
        \caption{Task conditions (dots) along their first two PCA axes, colored using the spatial $\beta$-map correlation.}
        \label{fig:pca_maps}
    \end{subfigure}
    \caption{}
    \label{fig:semantic_tuning_combined}
\end{figure}

\paragraph{Generated trajectories preserve semantic structure in task space.}
Beyond individual contrasts, we test whether generations preserve the organization of task
  space. We embed each GLM item using its sensory and instruction captions, compute semantic PCA
  axes, and compare real and synthetic coefficient maps obtained by regressing \(\beta\) values onto
  these axes. The resulting maps agree across real and generated data, and visualizing the first two
  semantic axes shows that \(f_{\text{spat}}\) maintains broader coverage where \(f_{\text{txt}}\)
  degrades (Figure~\ref{fig:semantic_tuning_combined}; Appendix \ref{app:task_space_pca}).

\paragraph{Compositional language for generation.}

Finally, we investigate the effect of dropping language-conditioning components on the synth/real $\beta$ correlation (Table \ref{tab:drop_hierarchy_compact}). Each ablation significantly decreases performance, with stronger degradation when multiple conditioning components are removed. This suggests that the model uses the conditioning signal compositionally, while also exhibiting partial redundancy between components. We observe substantial variation in degradation strength across contrasts, which we visualize in Appendix \ref{app:txt_token_dropout}.

\begin{wraptable}{r}{0.40\textwidth}  
\vspace{-\baselineskip}                
\centering
\footnotesize
\setlength{\tabcolsep}{4pt}
\begin{threeparttable}
  \caption{Synth/real $\beta$ correlations under hierarchical component dropping of language token $\mathbf{c}^\text{txt}$.}
  \label{tab:drop_hierarchy_compact}
  \begin{tabular}{@{}clc@{}}
    \toprule
    $|C|$ & Dropped components $C$ & $r$ \\
    \midrule
    0 & $\varnothing$ (PINC-Flow-$f_\text{txt}$) & \textbf{.440} \\
    \midrule
    1 & $r$ (response)    & .410$^{*}$ \\
      & $i$ (instruction) & .397$^{*}$ \\
      & $s$ (sensory)     & .395$^{*}$ \\
    \midrule
    2 & $i+r$ & .350$^{*}$ \\
      & $s+r$ & .355$^{*}$ \\
      & $i+s$ & .229$^{*}$ \\
    \midrule
    3 & $i+s+r$ \;(\emph{uncond.}) & .023$^{*}$ \\
    \bottomrule
  \end{tabular}
  \begin{tablenotes}[flushleft]
  \scriptsize
  \item $n=281$. All 12 paired one-sided $t$-tests over tasks $^{*}p_{\mathrm{Bonf}}<10^{-4}$.
  \end{tablenotes}
\end{threeparttable}
\vspace{-\baselineskip}                
\end{wraptable}

\section{Conclusion and Limitations}

We introduced PINC-Flow, a per-timestep conditioned diffusion transformer for generating whole-cortex fMRI dynamics under unseen cognitive tasks. Across held-out task conditions, the model recovers task-evoked activation patterns from generated trajectories, preserves region-specific recruitment across task space, and supports counterfactual changes to event timing and composition. This type of model may in the future support (i) screening candidate designs (e.g. amount of task blocks; see Appendix \ref{app:design_example} for an example), (ii) the generation of subject-conditioned synthetic data to validate an analysis pipeline and perform power estimation, and (iii) testing whether a hypothesis yields detectable effects given an experimental design.     

Some limitations deserve mention. First, our model extrapolates around the task-fMRI manifold covered by the training datasets, rather than to arbitrary cognitive experiments. Although HCP and IBC provide broad task coverage, far-extrapolative text conditions remain difficult and performance degrades with distance from the training support.  Further, although we find that $f_\text{spat}$ improves on a prior map, it's dependent on the quality of the map, which causes a weak predicted prior to underperform conditioning on language alone. Next, our splicing pretraining strategy creates only approximate counterfactuals that retain carryover from the original preceding event. Finally, language conditioning underspecifies fine-grained perceptual content, especially in low-level sensory cortex. Hybrid conditioning with modality-specific encoders is a natural extension, while our results indicate that the current language pathway remains well matched to whole-cortex cognitive dynamics.

\WFclear
\par\vspace{1em}   

\section{Acknowledgements and Disclosures}

This research was funded by Gemeinnützigen Hertie-Stiftung and the Deutsche Forschungsgemein-schaft (DFG) through FOR 5187 (project number 442075332). Additional support was provided by the Machine Excellence Cluster and DFG through the Germany’s Excellence Strategy (EXC 2064 - project number 390727645) and the following projects: CRC 1404 (project number 414984028), TRR 265 (project number 402170461), and RU 5363 (project number 459422098).

We thank Moritz Seiler and Nicolas Münster for their valuable comments on an early version of this manuscript. Finally, we are grateful to all dataset providers for access (UKB application number: 25163).

The authors declare no competing interests.

\bibliographystyle{plainnat}
\bibliography{references}

\appendix

\section{Dataset Details}
  \label{app:data_details}

We combine four datasets that differ in subject count, task diversity, and recording structure. The Human Connectome Project (HCP; \cite{barch2013function}) provides approximately \(1{,}100\) subjects across seven tasks, supporting held-out-subject evaluation. The Individual Brain Charting dataset (IBC; \citep{ponce2026individual}) includes 12 subjects who each perform 53 tasks, providing broad task-space coverage with at least one dedicated recording run per task and subject. The Nakai dataset \citep{nakai2020quantitative} comprises 6 subjects across 103 task conditions. Because Nakai task conditions alternate rapidly within each run across 18 runs per subject, we use it only for training to avoid biased task-level evaluation through temporal leakage. UK Biobank (UKB; \cite{miller2016multimodal}) contributes approximately \(39{,}000\) subjects of resting-state data, used unconditionally during training to improve modeling of intrinsic brain activity.

We match the sampling rate to 1\,s across datasets. Since the slowest original repetition time is 2\,s, we apply a 0.25\,Hz low-pass filter before resampling and z-score each parcel. Overall, the combined dataset contains 3.6M task-fMRI timepoints and 13.9M resting-state timepoints. Due to dataset imbalance, batches are sampled with non-uniform dataset weights to preserve gradient pressure on task data:
\[
w_\text{HCP}=1.0,\quad
w_\text{IBC}=0.85,\quad
w_\text{Nakai}=0.15,\quad
w_\text{UKB}=0.2.
\]

\textbf{Access.} Data access is governed by the original
providers of each dataset and can be requested through the
procedures described in the respective dataset citations. All
datasets require a data-use or data-sharing agreement, and all
have been widely used in prior neuroimaging and machine learning research.

 \section{Direct Text-to-Contrast Baseline}
  \label{app:direct_baseline}

  The Direct baseline predicts static contrast maps from the same language-derived task annotations used by the dynamics model. It decomposes each timestep's conditioning into an event component and, when applicable, a response component. Event components \(\hat{\boldsymbol{\beta}}^{\mathrm{event}}\) are predicted from the concatenated sensory and instruction embeddings using ridge regression. Response components \(\hat{\boldsymbol{\beta}}^{\mathrm{resp}}\) are predicted by nearest-neighbor lookup over the training fold, reflecting the small and discrete response space. When both event and response components are present, the predicted map is their sum; otherwise, the available component is used directly.

  Hyperparameters are selected by fold-internal cross-validation, with no held-out task leakage. We search ridge penalties
  \[
  \alpha \in \{1, 10, 100, 10^3, 10^4, 10^5\}
  \]
  and response nearest-neighbor values
  \[
  k \in \{1, 5, 10, 20\}.
  \]
  Model selection uses Pearson correlation between predicted and held-out training-fold contrast maps. We found that strong ridge regularization is typically selected, typically \(\alpha=10^4\), which improves spatial pattern recovery but shrinks prediction magnitudes. To restore realistic amplitudes without using held-out tasks, we rescale predictions within each Yeo-17 network to match the training-fold contrast norms. These networks by \cite{yeo2011organization} are widely-used in the neuroscience literature. The resulting Direct maps are used both as a baseline in contrast-recovery analyses and as optional predicted spatial priors for the dynamics model.

\section{Generative Baselines}
\label{app:generative_baselines}

\paragraph{HRF-VAR.} Given the held-out task schedule $s_k(t)$ and map $\hat{\boldsymbol{\beta}}_k$,
\begin{equation}
\hat{\mathbf x}_t = \sum_k \hat{\boldsymbol\beta}_k\,(\mathrm{hrf}*s_k)(t) + \mathbf U\boldsymbol\eta_t,\qquad
\boldsymbol\eta_t = \mathbf A\boldsymbol\eta_{t-1} + \boldsymbol\xi_t,\ \ \boldsymbol\xi_t\sim\mathcal N(\mathbf 0,\boldsymbol\Sigma),
\end{equation}
where $\mathbf{U}$ denote PCA loadings and $(\mathbf{A}, \boldsymbol\Sigma)$ are the VAR(1) parameters fit on the GLM residuals. Number of PCA components was selected via cross-validation from $\{10,20,30,40,50\}$, with a post-hoc analysis showing that the maximum was selected almost always. On average, fits were modestly better going from 40 to 50 components, indicating that more components might have yielded slightly better results.

\paragraph{Text-FIR.} We use the same caption embedding $\mathbf{e_t}$ as used for PINC-Flow and fit using ridge regression:
\begin{equation}
\hat{\mathbf x}_t = \sum_{\ell=0}^{L-1}\mathbf W_\ell\,\mathbf e_{t-\ell} + \mathbf b,
\end{equation}
The regularization strength was selected from $\{10^2, 10^3, 10^4, 10^5, 10^6\}$, with $10^5$ being selected for every fold.

\section{Additional Model and Optimization Hyperparameters}
\label{app:hyperparameters}

\begin{table}[H]
  \centering
  \caption{Summary of model architecture, dropout probabilities, and optimization hyperparameters.}
  \label{tab:hyperparams}
  \begin{tabular}{@{}ll@{}}
    \toprule
    \textbf{Hyperparameter} & \textbf{Value} \\
    \midrule
    \multicolumn{2}{@{}l}{\textit{Architecture \& Dimensions}} \\
    Embedding dimension & 1024 \\
    Number of heads & 16 \\
    Number of layers & 12 \\
    MLP ratio & 4.0 \\
    Dropout & 0.05 \\
    Context frames & 8 \\
    Generation frames & 16 \\
    Max context length & 24 \\
    \midrule
    \multicolumn{2}{@{}l}{\textit{Dropout Probabilities ($p_{\text{drop}}$)}} \\
    Spatial prior & 0.5 \\
    Context volumes & 0.2 \\
    Condition context\textsuperscript{a} & 0.15 \\
    Condition future\textsuperscript{b} & 0.1 \\
    Instruction embedding & 0.1 \\
    Sensory embedding & 0.1 \\
    Response embedding & 0.1 \\
    \midrule
    \multicolumn{2}{@{}l}{\textit{Training \& Optimization}} \\
    Batch size & 256 \\
    Number of epochs & 1000 (100 steps/epoch) \\
    Mixed precision (\texttt{use\_amp}) & True \\
    Gradient accumulation steps & 1 \\
    Optimizer & AdamW \\
    Learning rate & $2.4 \times 10^{-4}$ \\
    Weight decay & 0.05 \\
    AdamW betas & $(0.9, 0.999)$ \\
    Learning rate schedule & Cosine decay with warmup\textsuperscript{c} \\
    Warmup epochs & 10 \\
    Minimum LR ratio & 0.3 \\
    Cosine epochs & 500 \\
    \midrule
    \multicolumn{2}{@{}l}{\textit{Text Embedding Model}} \\
    Model & BAAI/bge-large-en-v1.5 \citep{bge_embedding} \\ 
    Dimensionality & 1024 \\
    \bottomrule
  \end{tabular}
  
  \vspace{0.5em}
  \raggedright
  \footnotesize
  \textsuperscript{a} All task conditioning for context volumes.\\
  \textsuperscript{b} All task conditioning for generation volumes.\\
  \textsuperscript{c} After warmup, cosine decay over \texttt{cosine\_epochs}, then constant at $\texttt{min\_lr\_ratio} \times \texttt{lr}$.
\end{table}

\section{Flow matching parameterization}
\label{app:fm_parameterization}

In the flow matching paradigm, models are commonly trained to predict the velocity. With the linear path $z_\tau = (1-\tau)z_0 + \tau z_1$, the velocity is $v = (z_1 - z_\tau)/(1-\tau)$. We here use an alternative to this "v-pred" strategy, which is to predict the endpoint ("x-pred") using the loss $\vert \hat{z}_1 - z_1 \vert^2$. We note that, via substitution: 
$$\vert \hat{z}_1 - z_1\vert^2 = (1 - \tau)^2 \vert \hat{v} - v \vert ^2$$
Therefore, training with "x-pred" is the same as training on the velocity loss with weight $w(\tau) = (1-\tau)^2$, yielding the same optimum but differing in the contribution of each point along the path. A priori, our motivation was that the endpoint target is a clean data sample at a fixed scale. At high $\tau$,  $z_1$ is almost entirely determined by $z_\tau$ and almost not at all by $z_0$. As such, the velocity target $z_1 - z_0$ contains irreducible conditional variance while the endpoint target does not. We run a post-hoc ablation, but find there to be a minimal and likely neglible difference between the two parameterizations. Here we show v-pred minus x-pred:

\begin{table}[h]
\centering
\begin{tabular}{lcc}
\hline
Metric & Item-weighted $\Delta$ & Task-macro $\Delta$ \\
\hline
Contrasts & $-.020$ & $\sim 0.000$ \\
Single conditions & $-.041$ & $-.029$ \\
\hline
\end{tabular}
\end{table}

\section{Splicing ablation}
\label{app:splicing_ablation}

We retrained models without splicing. While OOD performance is unchanged, we find that transitions involving in-distribution tasks are degraded. Specifically, we condition models using contexts containing task block $A$ while requesting $C$, whereas $B$ was the observed successor during training. We find that splicing modestly improves recovery of $C$ ($r=.74 \rightarrow .81$) and removes reversion to the context-implied $B$ ($r=.17 \rightarrow -.02$) (Table \ref{tab:splice_ablation}).

\begin{table}[h]
\centering
\small
\caption{Transition-splicing ablation (PINC-Flow-$f_\text{txt}$). We test in-distribution (ID), where the context seen during training implies block B will follow, but block C is requested instead (paired leave-one-subject-out CI). For OOD, we show map recovery over all seven held-out-task folds (mean$\pm$std over items; CI: bootstrap over 52 tasks). Note that results differ minimally from the main results: for computational efficiency we only evaluate on one (rather than all six) captions.}
\label{tab:splice_ablation}
\begin{tabular}{@{}lccc@{}}
  \toprule
  & No splice & Splice & $\Delta$ [95\% CI] \\
  \midrule
  \multicolumn{4}{@{}l}{\emph{ID: counterfactual (MOTOR)}} \\
  $r$ w/ requested C $\uparrow$  & .742 & \textbf{.811} & $+$.069 [.052, .085] \\
  $r$ w/ obsolete B $\downarrow$ & .166 & \textbf{$-$.025} & $-$.191 [$-$.212, $-$.171] \\
  \midrule
  \multicolumn{4}{@{}l}{\emph{OOD: map recovery}} \\
  $r$ (all items) & .433 $\pm$ .308 & .423 $\pm$ .315 & $-$.010 [$-$.034, .012] \\
  \bottomrule
\end{tabular}
\end{table}

\section{Compute Usage}
\label{app:compute_usage}

Training our model for 100K steps takes 1.5 days on a single L40 GPU (48GB). Across seven folds, this results in an aggregate 10.5 days. 

While evaluation costs themselves are negligible, they require model inference for synthetic data generation. This takes approximately three hours per fold on a 20GB GPU. The main results compare five models across 7 folds, yielding an approximate 105 hour compute usage. We ablate two models under 4 guidance scales and one model under 7 dropout settings, yielding another 315 hours of 20GB GPU usage. 

In total, the experiments reported in the paper therefore use about 672 GPU hours. We estimate preliminary work to have consumed twice this amount at approximately 1400 GPU hours.

\section{Stochastic Sampling}
  \label{app:stochastic_fc}

All main task-generation results use deterministic Euler sampling. We additionally evaluate a
stochastic sampler to test whether context-related over-conditioning can be reduced without
changing the trained model. At each Euler step except the final step, we add temporally
correlated AR(1) Gaussian noise in latent space with autocorrelation \(\rho=0.7\) and scale
\(\varepsilon=0.6\), scaled by \(\sqrt{\Delta\tau}(1-\tau)\) so that stochasticity decreases as the sampler approaches $\tau=1$. 

Deterministic sampling produces realistic task contrasts and basic time-series statistics, but generated FC can be more similar to the conditioning context than expected from real run-to-run FC. Stochastic sampling improves calibration, reducing generated-context FC correlations toward the real reference distribution while preserving the expected monotonic effect of context guidance.

\begin{figure}[h]
    \centering
    \includegraphics[width=1\linewidth]{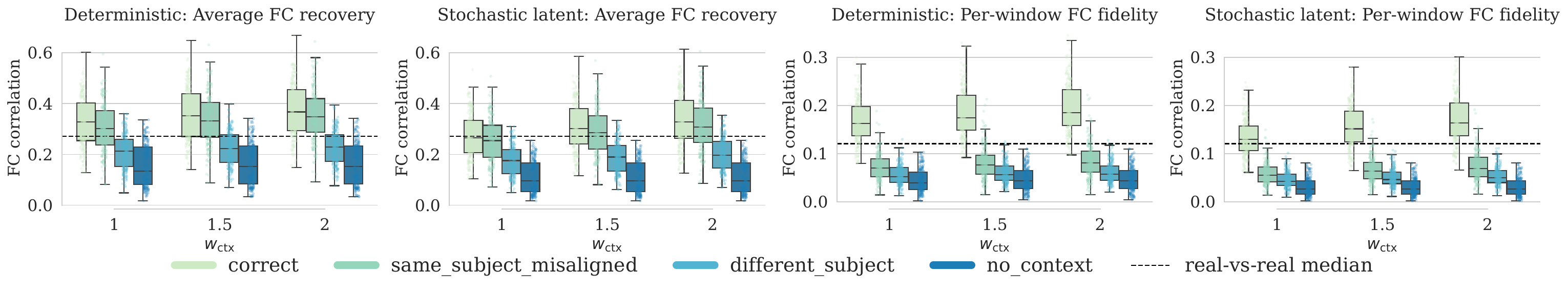}
    \caption{Functional connectivity in generated trajectories can be modulated via guidance and is better calibrated using stochastic sampling. Mean$\pm$std across subjects (left) and windows (right).}
    \label{fig:fc}
\end{figure}

\section{Design reliability screening example}
\label{app:design_example}

\begin{figure}[h]
    \centering
    \includegraphics[width=1\linewidth]{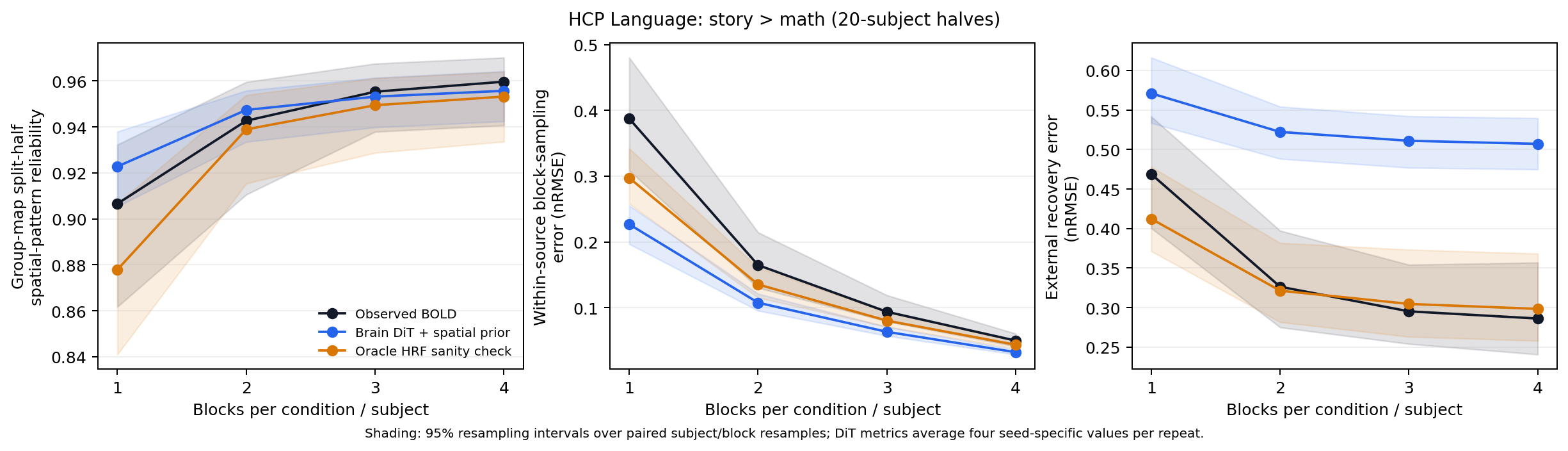}
    \caption{Using the HCP Language task, we investigate the design screening capability of PINC-Flow.}
    \label{fig:design_example}
\end{figure}

As an example future use-case for conditional generative models, we investigate the model for design screening (Figure \ref{fig:design_example}). Concretely, for the HCP language task, we generate between one and four task blocks and compute the "story > math" map. On the left, we plot the split-half reliability across the number of task blocks. We find that our DiT model (using an independent cross-run prior) does well and captures a stable group map after 2-3 blocks. In the middle, we measure within-source sampling error, here PINC-Flow again reproduces the shape of the noise-vs-blocks curve, although it underestimates the noise compared to real data. This would lead to an underestimating of how many blocks are needed. Finally, on the right, the error with respect to a real map shows that our model has a bias-based error for this particular OOD task, as error does not significantly reduce with more data. These post-hoc analyses were performed with default sampling parameters and we leave attempts to improve this type of calibration via training and inference tuning for future work.

\section{Task Space PCA}
\label{app:task_space_pca}

Following PCA of the concatenated sensory and instruction text embeddings, we investigated the resulting low-dimensional semantic axes. To interpret these axes, we projected all experimental events onto the first five PCs. We then examined the items (comprising single conditions and task contrasts) with the high positive and negative dot product scores for each PC. Top and bottom four scores are presented in Table \ref{tab:super-long-table}.
This analysis revealed clear semantic distinctions for four of the five axes. For instance, PC 1 discriminates between tasks involving language processing (e.g., “auditory theory of mind,” “audio sentence,” “listen to the story,” “general auditory,” “audio story about false beliefs”) and motor execution (e.g., “move right/left hand,” “move right/left foot”). 
PC 2 differentiates visual evaluation tasks (e.g., gamble presentation, preferences for faces/paintings/houses/food) from auditory stimulation (e.g., listening to different sounds).
PC 3 separates visual encoding (e.g., dots/letters stimuli) from motor tasks. 
PC 5 contrasts embodied processes (e.g., moving tongue/hand, face preference, emotion recognition) with mental simulation tasks (e.g., mental time travel west-east/north-south).
However, we did not find a clearly identifiable semantic boundary in PC 4, therefore it was excluded from the visualization.

{\tiny
\begin{longtable}{crrclcl}
\caption{Semantic PCA axes 1-5 - conditions and contrasts text representations projected onto the respective PC axes and sorted by the resulted dot product (score). We present the first and last 4 values. Item kinds are abbreviated as s.\ (single condition) and c.\ (contrast).} \\
\toprule
axis & rank & score & dataset & task & kind & item name \\
\midrule
\endfirsthead

\caption[]{Semantic PCA axes 1-5 (continued).} \\
\toprule
axis & rank & score & dataset & task & kind & item name \\
\midrule
\endhead

\bottomrule
\endfoot

1 & 1 & \textcolor{red}{2.355} & IBC & MathLanguage & s. & events:theory\_of\_mind\_auditory \\
 & 2 & \textcolor{red}{2.332} & IBC & ArchiStandard & s. & events:audio\_sentence \\
 & 3 & \textcolor{red}{2.293} & HCP & LANGUAGE & s. & story\_listen \\
 & 4 & \textcolor{red}{2.276} & IBC & MathLanguage & s. & events:general\_auditory \\
 & \vdots & \vdots & \vdots & \vdots & \vdots & \vdots \\
 & -4 & \textcolor{blue}{-1.588} & HCP & MOTOR & c. & move\_rf\_gt\_fixation \\
 & -3 & \textcolor{blue}{-1.588} & HCP & MOTOR & s. & move\_rf \\
 & -2 & \textcolor{blue}{-1.612} & HCP & MOTOR & c. & move\_rh\_gt\_fixation \\
 & -1 & \textcolor{blue}{-1.612} & HCP & MOTOR & s. & move\_rh \\
\midrule
2 & 1 & \textcolor{red}{1.905} & IBC & NARPS & s. & events:narps\_gamble\_gain\_hi\_loss\_hi \\
 & 2 & \textcolor{red}{1.902} & IBC & NARPS & s. & events:narps\_gamble\_gain\_lo\_loss\_hi \\
 & 3 & \textcolor{red}{1.901} & IBC & NARPS & c. & stim\_gt\_fixation \\
 & 4 & \textcolor{red}{1.900} & IBC & NARPS & s. & events:narps\_gamble\_gain\_hi\_loss\_lo \\
 & \vdots & \vdots & \vdots & \vdots & \vdots & \vdots \\
 & -4 & \textcolor{blue}{-2.647} & IBC & Audi & s. & events:french\_speech \\
 & -3 & \textcolor{blue}{-2.672} & IBC & Audi & s. & events:alphabet\_audio \\
 & -2 & \textcolor{blue}{-2.678} & IBC & Audi & s. & events:coughing\_sounds \\
 & -1 & \textcolor{blue}{-2.687} & IBC & Audi & c. & all\_audio\_gt\_silence \\
\midrule
3 & 1 & \textcolor{red}{1.918} & IBC & MVIS & s. & events:six\_dots\_array \\
 & 2 & \textcolor{red}{1.851} & IBC & MVIS & s. & events:four\_dots\_array \\
 & 3 & \textcolor{red}{1.846} & IBC & MVEB & s. & events:four\_letters\_same\_string \\
 & 4 & \textcolor{red}{1.811} & IBC & MVEB & c. & string\_encoding\_gt\_fixation \\
 & \vdots & \vdots & \vdots & \vdots & \vdots & \vdots \\
 & -4 & \textcolor{blue}{-2.176} & HCP & MOTOR & s. & move\_rf \\
 & -3 & \textcolor{blue}{-2.205} & HCP & MOTOR & s. & move\_rh \\
 & -2 & \textcolor{blue}{-2.205} & HCP & MOTOR & c. & move\_rh\_gt\_fixation \\
 & -1 & \textcolor{blue}{-2.230} & IBC & MTTWE & s. & events:westside\_event\_audio \\
\midrule
4 & 1 & \textcolor{red}{1.946} & IBC & MVEB & s. & events:two\_letters\_different\_string \\
 & 2 & \textcolor{red}{1.934} & IBC & MVEB & c. & string\_encoding\_gt\_fixation \\
 & 3 & \textcolor{red}{1.924} & IBC & MVEB & s. & events:two\_letters\_same\_string \\
 & 4 & \textcolor{red}{1.900} & IBC & MVEB & s. & events:four\_letters\_different\_string \\
 & \vdots & \vdots & \vdots & \vdots & \vdots & \vdots \\
 & -4 & \textcolor{blue}{-2.717} & IBC & BiologicalMotion2 & s. & events:modified\_inverted \\
 & -3 & \textcolor{blue}{-2.729} & IBC & BiologicalMotion2 & s. & events:natural\_upright \\
 & -2 & \textcolor{blue}{-2.768} & IBC & BiologicalMotion2 & c. & biological\_motion\_gt\_fixation \\
 & -1 & \textcolor{blue}{-2.799} & IBC & BiologicalMotion2 & s. & events:natural\_inverted \\
\midrule
5 & 1 & \textcolor{red}{1.894} & IBC & Moto & s. & events:tongue\_left \\
 & 2 & \textcolor{red}{1.884} & IBC & Moto & s. & events:hand\_left \\
 & 3 & \textcolor{red}{1.848} & IBC & Moto & s. & events:tongue\_right \\
 & 4 & \textcolor{red}{1.815} & IBC & Moto & c. & saccade\_gt\_other\_movements \\
 & \vdots & \vdots & \vdots & \vdots & \vdots & \vdots \\
 & -4 & \textcolor{blue}{-3.455} & IBC & MTTNS & s. & events:southside\_close\_event \\
 & -3 & \textcolor{blue}{-3.478} & IBC & MTTNS & c. & all\_events\_gt\_fixation \\
 & -2 & \textcolor{blue}{-3.478} & IBC & MTTNS & c. & space\_event\_gt\_time\_event \\
 & -1 & \textcolor{blue}{-3.483} & IBC & MTTNS & s. & events:northside\_close\_event \\
\label{tab:super-long-table}
\end{longtable}
}

\section{Language-Token component dropout - extra results}
  \label{app:txt_token_dropout}

\begin{figure}[H]
  \centering
  \includegraphics[width=1\linewidth]{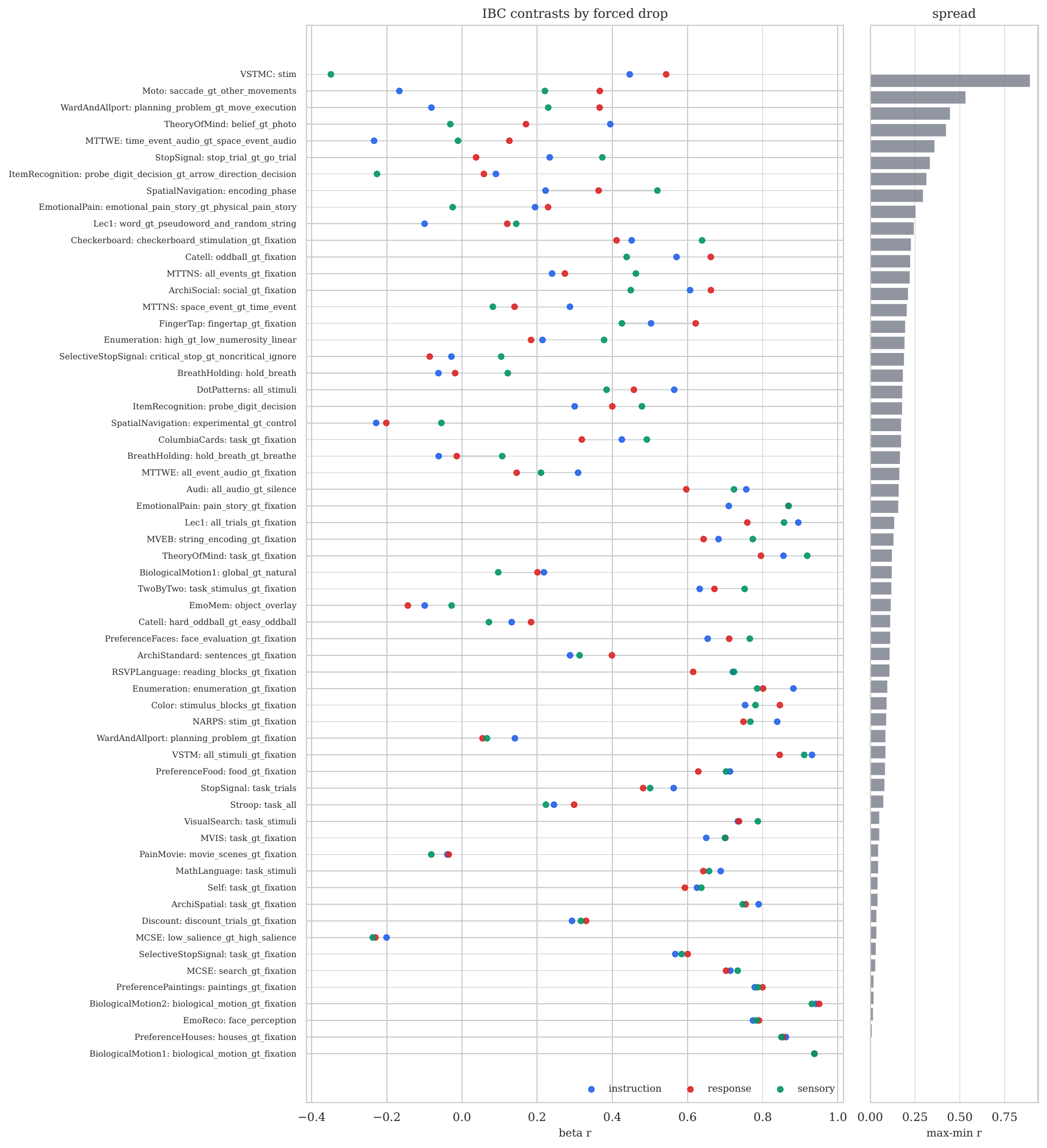}
  \caption{We show the effects on synth/real $\beta$ correlations for individual IBC contrasts when dropping out a single language token components during OOD evaluation.}
  \label{fig:IBC_contrast_spread}
\end{figure}

\section{Counterfactual Training details}
\label{app:counterfactual_training}
fMRI runs use fixed condition orderings, so the model repeatedly observes the same condition-to-condition transitions and tends to overfit to them, which harms compositional generalization. To diversify transitions, with probability $p_{\text{cf}}$ we splice two discontinuous segments from the same subject, creating a novel condition transition with a temporal discontinuity at splice point $t_s$. 

We therefore weight the per-timestep loss for $t > t_s$ by

\[
\tilde w_t =
\begin{cases}
\dfrac{hrf(t - t_s)}
{hrf(\tau_{\mathrm{peak}})},
& t_s \leq t \leq t_s + \tau_{\mathrm{peak}}, \\[8pt]
1,
& t_s + \tau_{\mathrm{peak}} < t \leq T_g,
\end{cases}
\qquad
w_t =
\frac{T_g}{\sum_{u=1}^{T_g} \tilde w_u}\,\tilde w_t .
\]
Here $hrf$ is the canonical double-gamma HRF and $\tau_{\mathrm{peak}} \approx 5\,\mathrm{s}$ is the location of its peak. The unnormalized weight $\tilde w_t$ is zero at the splice point, where the model cannot reconstruct the discontinuity, and recovers smoothly over the HRF rise time. We rescale to $w_t$ so that $\sum_{t=1}^{T_g} w_t = T_g$, preserving the average loss scale over the generation window across spliced and unspliced samples.

\section{Counterfactual interpretation and splice approximation}
\label{app:counterfactual_interp}

We use the term counterfactual to mean the model-implied response under a task composition, transition, or timing not observed during training. For manipulations covered by the training data, the generated distribution can be seen as an interventional distribution under consistency, no interference, and positivity. For held-out tasks, the model must instead extrapolate from components seen during training. Finally, conditional flows do not identify how an individual's response for a given task pairs with their response for another task. As such, we claim we perform counterfactual simulation, rather than identified individual potential outcomes.

\paragraph{Interventional reading.} Let $c$ denote a task specification controlled by an experimenter. Under consitency (the observed response equals the potential response under the realized $c$), no interference (one subject's response is unaffected by another's assignment), and positivity (each $c$ of interest has support in training), $p_1(z_1 \vert c^\text{txt}, c^\text{ctx})$ coincides with the distribution $p(z_1 \vert \text{do}(c), c^\text{ctx})$. For held-out tasks, positivity fails as generation extrapolates over components.

\paragraph{Splice decomposition.} Let $\mathbf{x}_B (t \vert A)$ denote the response to event $B$ following the event $A$, $\mathbf{x}_B(t \vert A) = \mathbf{r}_B (t \vert A) + \mathbf{c}_A(t) + \varepsilon_t$, where $\mathbf{r}_B$ is the event-related respones to $B$, $\mathbf{c}_A$ the residual activity bleeding from $A$, and $\varepsilon_t$ noise. Splicing then resuses a segment observed after $A$ in place of the unobserved response after $C$:
$$
\mathbf{x}_B(t\mid A) - \mathbf{x}_B(t\mid C) = \underbrace{\mathbf{r}_B(t\mid A) - \mathbf{r}_B(t\mid C)}_{\text{history-dependent modulation}} + \underbrace{\mathbf{c}_A(t) - \mathbf{c}_C(t)}_{\text{predecessor carryover}} + (\boldsymbol\varepsilon_t - \boldsymbol\varepsilon'_t).
$$
At the $\beta$-map level, the splice is an approxiamte counterfactual target when $B$'s spatial response is locally stable across precedessors. For the timeseries level however, carryover decays approximately over one HRF, which motivates the HRF-based loss weighting.

\paragraph{Non-identifiability of pairings.} Conditional flows match the task-specific marginals $p(z_1 \vert c)$. Different latent parameterizations can reproduce the same marginals while yielding different pairings of an individual's responses cross-task. The generated cross-task pairs are therefore only 'model-implied', rather than identified potential outcomes.

\section{Extended related work}
\label{app:extra_related_work}

\textbf{Encoding or generative models.} TRIBEv2 \citep{d2026foundation} is a stimulus-encoding model, as it maps audio, video, and text streams to a deterministic, group-level BOLD trajectory. Meanwhile, PINC-Flow conditions on a factorized, textual experimental description, subject context, and an optional spatial prior to sample a distribution over recordings. We incude motor responses and task instructions to also model active rather than only perceptual paradigms. Conversely, the raw-stimulus encoding used by TRIBEv2 is better situated to preserve perceptual details missed by language descriptions. BrainDiVE \citep{luo2023brain} and BeNeDiff \citep{lee2023exploring} are methods using diffusion in the opposite direction. As they generate images or behaviour videos associated with brain factors, they ask what neural factors represents, while we ask what neural dynamics follow from an experimental condition. We provide a visual comparison across these methods in Figure \ref{fig:comparison}.

\begin{figure}[H]
  \centering
\begin{tikzpicture}[font=\sffamily\scriptsize]
\def\lab{0}        
\def\cA{6.3}       
\def\cB{8.0}       
\def\cC{9.8}       
\def\cD{11.6}      
\def\dy{0.42}      
\fill[grp!12,rounded corners=2pt] (\cA-0.8,0.95) rectangle (\cB+0.8,1.40);
\node[grp!80!black] at ($(\cA,1.175)!0.5!(\cB,1.175)$) {neural activity $\rightarrow$ content};
\fill[ours!12,rounded corners=2pt] (\cC-0.8,0.95) rectangle (\cD+0.8,1.40);
\node[ours!80!black] at ($(\cC,1.175)!0.5!(\cD,1.175)$) {experiment $\rightarrow$ neural activity};
\node[align=center] at (\cA,0.55) {\textbf{BrainDiVE}};
\node[align=center] at (\cB,0.55) {\textbf{BeNeDiff}};
\node[align=center] at (\cC,0.55) {\textbf{TRIBE v2}};
\node[align=center,text=ours!85!black] at (\cD,0.55) {\textbf{PINC-Flow}\\(ours)};
\draw[black!60,line width=0.5pt] (\lab,0.05) -- (\cD+0.85,0.15);
\fill[ours!7] (\cD-0.8,0.03) rectangle (\cD+0.8,{0.15-13*\dy-0.12});
\foreach \i/\txt/\a/\b/\c/\d in {
  1/{Generated output}/{images}/{behavior video}/{fMRI}/{fMRI},
  2/{Conditioning signal}/{ROI target}/{neural latents}/{raw stimuli}/{captions},
  3/{Recording modality}/{human fMRI}/{mouse Ca\textsuperscript{2+}}/{human fMRI}/{human fMRI}
}{
  \node[anchor=west] at (\lab,{-\i*\dy+0.1}) {\txt};
  \node at (\cA,{-\i*\dy+0.1}) {\textcolor{black!70}{\a}};
  \node at (\cB,{-\i*\dy+0.1}) {\textcolor{black!70}{\b}};
  \node at (\cC,{-\i*\dy+0.1}) {\textcolor{black!70}{\c}};
  \node at (\cD,{-\i*\dy+0.1}) {\textcolor{black!70}{\d}};
}
\draw[black!25] (\lab,{-3.5*\dy+0.1}) -- (\cD+0.85,{-3.5*\dy+0.1});
\foreach \i/\txt/\a/\b/\c/\d in {
  4/{Generates neural time series}/\none/\none/\full/\full,
  5/{Samples a distribution}/\full/\full/\none/\full,
  6/{Raw stimulus input}/\na/\na/\full/\none,
  7/{Instructions \& motor responses as inputs}/\na/\na/\none/\full,
  8/{Per-timestep control of event timing}/\na/\na/\full/\full,
  9/{Steerable by a spatial / neural hypothesis}/\full/\full/\none/\full,
  10/{Individualized output for new subjects w/o retraining}/\none/\na/\half/\full,
  11/{Novel stimuli at inference}/\na/\na/\full/\half,
  12/{Held-out task paradigms}/\na/\na/\full/\full,
  13/{Subcortical targets}/\na/\na/\full/\none
}{
  \node[anchor=west] at (\lab,{-\i*\dy+0.1}) {\txt};
  \node at (\cA,{-\i*\dy+0.1}) {\a};
  \node at (\cB,{-\i*\dy+0.1}) {\b};
  \node at (\cC,{-\i*\dy+0.1}) {\c};
  \node at (\cD,{-\i*\dy+0.1}) {\d};
}
\draw[black!60,line width=0.5pt] (\lab,{-13.5*\dy+0.1}) -- (\cD+0.85,{-13.5*\dy+0.1});
\node[anchor=west,text=black!75] at (\lab,{-14.2*\dy+0.1})
  {\full~native \quad \half~partial / indirect \quad \none~not supported \quad \na~not applicable};
\end{tikzpicture}
  \caption{Methods comparison.}
  \label{fig:comparison}
\end{figure}

\section{Synthetic Run Construction and GLM Evaluation}
  \label{app:glm_evaluation}

  \paragraph{Windowed generation.}
  The dynamics model generates fixed-length windows rather than full runs in a single forward process. For evaluation, we therefore construct synthetic counterparts of held-out task recordings by
  repeatedly generating non-overlapping windows of length \(T_g\). Each generated window receives the task annotations for the corresponding target timesteps and leakage-controlled context volumes. The generated windows are then concatenated into a synthetic run-like sequence and analyzed with the same GLM machinery as the corresponding real data.

  For HCP, generation windows begin at the start of each task block. This covers nearly all task-relevant volumes, although short inter-block or trailing periods may not be represented. HCP also includes
  pre-task and fixation periods, allowing context volumes to be drawn from periods whose task conditioning can be truthfully specified as fixation. For IBC, which does not provide the same block
  structure, we generate non-overlapping windows along the target recording. Because IBC provides little pre-task signal, context is drawn from a different task recording from the same subject; task-
  conditioning tokens are masked for these context timesteps to prevent leakage. Across datasets, task conditioning is supplied only for generated volumes.

  \paragraph{Matched real and synthetic GLMs.}
  Although the generative model receives sensory, instruction, and response captions at every timestep, evaluation follows standard task-fMRI practice and uses event-level GLMs. Each event regressor
  encodes the event's duration or fractional occupancy within each volume and is convolved with the canonical HRF. Thus, a brief stimulus occupying half of a 1-second volume contributes weight \(0.5\),
  whereas sustained events contribute positive weights, typically \(1\), across multiple successive volumes.

  For each held-out condition or contrast, we fit identical design matrices to matched real and synthetic samples. This yields a real contrast map \(\boldsymbol{\beta}_k\) and a synthetic contrast map
  \(\hat{\boldsymbol{\beta}}_k\), which are compared across parcels. Real and synthetic analyses are always matched in sample size and event coverage, even when the dataset-specific windowing strategy
  does not cover every volume of the original run. For brevity, we use ``contrast'' to refer both to single event maps and to differences between event maps.

  \subsection{Contrast Inventory and Filtering}
  \label{app:contrast_inventory}

  \paragraph{Filtering.}
  The GLM inventory below includes only contrasts that passed a reliability check. Specifically, IBC contrasts with split-half reliability below \(r=0.25\) were
  excluded.

\paragraph{HCP contrasts.}
For HCP, contrasts are defined per task as follows:
\begin{description}
    \item[EMOTION.] face-shape perceptual blocks, plus shape against fixation.
    \item[GAMBLING.] reward and loss feedback events in both directions.
    \item[LANGUAGE.] story listening against math listening.
    \item[MOTOR.] each movement block (left/right hand, left/right foot, tongue) against fixation.
    \item[RELATIONAL.] relational reasoning blocks against shape-matching control blocks.
    \item[SOCIAL.] mental-interaction movie blocks against random-motion movie blocks.
    \item[WM.] 2-back working-memory blocks against 0-back blocks.
\end{description}

\paragraph{IBC contrast approach.}
IBC contrasts are task-specific linear combinations of event and, where needed, response regressors. Broad task-vs-baseline contrasts are used for tasks with a clear overall activation target; narrower task-vs-task contrasts are used when the experiment has a natural within-task comparison. Multi-condition contrasts average related regressors with equal or prespecified weights, while fixation/rest/silence terms are treated as baseline terms rather than standalone single conditions.

\paragraph{HCP contrast definitions.}
{\tiny
\begin{description}
    \item[EMOTION.] Contrasts: \texttt{stim\_face\_gt\_stim\_shape} = \texttt{stim\_face} $-$ \texttt{stim\_shape}; \texttt{stim\_shape\_gt\_fixation} = \texttt{stim\_shape} $-$ \texttt{fixation}. Single conditions: \texttt{stim\_face}, \texttt{stim\_shape}.
    \item[GAMBLING.] Contrasts: \texttt{reward\_feedback\_gt\_loss\_feedback} = \texttt{reward\_feedback} $-$ \texttt{loss\_feedback}; \texttt{loss\_feedback\_gt\_reward\_feedback} = \texttt{loss\_feedback} $-$ \texttt{reward\_feedback}. Single conditions: \texttt{reward\_feedback}, \texttt{loss\_feedback}.
    \item[LANGUAGE.] Contrasts: \texttt{story\_listen\_gt\_math\_listen} = \texttt{story\_listen} $-$ \texttt{math\_listen}. Single conditions: \texttt{story\_listen}, \texttt{math\_listen}.
    \item[MOTOR.] Contrasts: \texttt{move\_lh\_gt\_fixation} = \texttt{move\_lh} $-$ \texttt{fixation}; \texttt{move\_rh\_gt\_fixation} = \texttt{move\_rh} $-$ \texttt{fixation}; \texttt{move\_lf\_gt\_fixation} = \texttt{move\_lf} $-$ \texttt{fixation}; \texttt{move\_rf\_gt\_fixation} = \texttt{move\_rf} $-$ \texttt{fixation}; \texttt{move\_t\_gt\_fixation} = \texttt{move\_t} $-$ \texttt{fixation}. Single conditions: \texttt{move\_lh}, \texttt{move\_rh}, \texttt{move\_lf}, \texttt{move\_rf}, \texttt{move\_t}.
    \item[RELATIONAL.] Contrasts: \texttt{block\_rel\_gt\_block\_match} = \texttt{block\_rel} $-$ \texttt{block\_match}. Single conditions: \texttt{block\_rel}, \texttt{block\_match}.
    \item[SOCIAL.] Contrasts: \texttt{movie\_mental\_gt\_movie\_random} = \texttt{movie\_mental} $-$ \texttt{movie\_random}. Single conditions: \texttt{movie\_mental}, \texttt{movie\_random}.
    \item[WM.] Contrasts: \texttt{block\_2bk\_gt\_block\_0bk} = \texttt{block\_2bk} $-$ \texttt{block\_0bk}. Single conditions: \texttt{block\_2bk}, \texttt{block\_0bk}.
\end{description}
}

HCP total: 13 contrasts and 17 single conditions.

\paragraph{IBC contrast definitions.}
{\tiny
\begin{description}
    \item[ArchiSocial.] Contrasts: \texttt{social\_gt\_fixation} = 0.25\,\texttt{events:false\_belief\_audio\_story} + 0.25\,\texttt{events:false\_belief\_video\_story} + 0.25\,\texttt{events:false\_belief\_why\_prompt} + 0.25\,\texttt{events:triangle\_mental}. Single conditions: \texttt{events:false\_belief\_audio\_story}, \texttt{events:false\_belief\_video\_story}, \texttt{events:false\_belief\_why\_prompt}, \texttt{events:triangle\_mental}.
    \item[ArchiSpatial.] Contrasts: \texttt{task\_gt\_fixation} = 0.2\,\texttt{events:object\_grasp} + 0.2\,\texttt{events:object\_orientation} + 0.2\,\texttt{events:rotation\_hand} + 0.2\,\texttt{events:rotation\_side} + 0.2\,\texttt{events:saccade} - \texttt{events:fixation}. Single conditions: \texttt{events:object\_grasp}, \texttt{events:object\_orientation}, \texttt{events:rotation\_hand}, \texttt{events:rotation\_side}, \texttt{events:saccade}.
    \item[ArchiStandard.] Contrasts: \texttt{sentences\_gt\_fixation} = 0.5\,\texttt{events:audio\_sentence} + 0.5\,\texttt{events:video\_sentence} - \texttt{events:fixation}. Single conditions: \texttt{events:audio\_sentence}, \texttt{events:video\_sentence}.
    \item[Audi.] Contrasts: \texttt{all\_audio\_gt\_silence} = 0.08333\,\texttt{events:alphabet\_audio} + 0.08333\,\texttt{events:animal\_sounds} + 0.08333\,\texttt{events:coughing\_sounds} + 0.08333\,\texttt{events:crying\_sounds} + 0.08333\,\texttt{events:environmental\_sounds} + 0.08333\,\texttt{events:french\_speech} + 0.08333\,\texttt{events:human\_non\_speech\_sounds} + 0.08333\,\texttt{events:laughing\_sounds} + 0.08333\,\texttt{events:music\_audio} + 0.08333\,\texttt{events:reversed\_speech} + 0.08333\,\texttt{events:suomi\_speech} + 0.08333\,\texttt{events:yawning\_sounds} - \texttt{events:silence\_baseline}. Single conditions: \texttt{events:alphabet\_audio}, \texttt{events:animal\_sounds}, \texttt{events:coughing\_sounds}, \texttt{events:crying\_sounds}, \texttt{events:environmental\_sounds}, \texttt{events:french\_speech}, \texttt{events:human\_non\_speech\_sounds}, \texttt{events:laughing\_sounds}, \texttt{events:music\_audio}, \texttt{events:reversed\_speech}, \texttt{events:suomi\_speech}, \texttt{events:yawning\_sounds}.
    \item[BiologicalMotion1.] Contrasts: \texttt{biological\_motion\_gt\_fixation} = 0.25\,\texttt{events:global\_inverted} + 0.25\,\texttt{events:global\_upright} + 0.25\,\texttt{events:natural\_inverted} + 0.25\,\texttt{events:natural\_upright} - \texttt{events:fixation}; \texttt{global\_gt\_natural} = 0.5\,\texttt{events:global\_inverted} + 0.5\,\texttt{events:global\_upright} - 0.5\,\texttt{events:natural\_inverted} - 0.5\,\texttt{events:natural\_upright}. Single conditions: \texttt{events:global\_inverted}, \texttt{events:global\_upright}, \texttt{events:natural\_inverted}, \texttt{events:natural\_upright}.
    \item[BiologicalMotion2.] Contrasts: \texttt{biological\_motion\_gt\_fixation} = 0.25\,\texttt{events:modified\_inverted} + 0.25\,\texttt{events:modified\_upright} + 0.25\,\texttt{events:natural\_inverted} + 0.25\,\texttt{events:natural\_upright} - \texttt{events:fixation}. Single conditions: \texttt{events:modified\_inverted}, \texttt{events:modified\_upright}, \texttt{events:natural\_inverted}, \texttt{events:natural\_upright}.
    \item[BreathHolding.] Contrasts: \texttt{hold\_breath} = \texttt{events:hold\_breath}; \texttt{hold\_breath\_gt\_breathe} = \texttt{events:hold\_breath} - \texttt{events:breathe}. Single conditions: \texttt{events:hold\_breath}, \texttt{events:breathe}.
    \item[Catell.] Contrasts: \texttt{oddball\_gt\_fixation} = 0.5\,\texttt{events:easy\_oddball} + 0.5\,\texttt{events:hard\_oddball} - \texttt{events:fixation}; \texttt{hard\_oddball\_gt\_easy\_oddball} = \texttt{events:hard\_oddball} - \texttt{events:easy\_oddball}. Single conditions: \texttt{events:easy\_oddball}, \texttt{events:hard\_oddball}.
    \item[Checkerboard.] Contrasts: \texttt{checkerboard\_stimulation\_gt\_fixation} = \texttt{events:checkerboard\_stimulation} - \texttt{events:fixation}. Single conditions: \texttt{events:checkerboard\_stimulation}.
    \item[Color.] Contrasts: \texttt{stimulus\_blocks\_gt\_fixation} = 0.5\,\texttt{events:achromatic\_block} + 0.5\,\texttt{events:chromatic\_block} - \texttt{events:fixation}. Single conditions: \texttt{events:achromatic\_block}, \texttt{events:chromatic\_block}.
    \item[ColumbiaCards.] Contrasts: \texttt{task\_gt\_fixation} = 0.3333\,\texttt{events:cards\_risk\_hi} + 0.3333\,\texttt{events:cards\_risk\_lo} + 0.3333\,\texttt{events:reward\_feedback} - \texttt{events:fixation}. Single conditions: \texttt{events:cards\_risk\_hi}, \texttt{events:cards\_risk\_lo}, \texttt{events:reward\_feedback}.
    \item[Discount.] Contrasts: \texttt{discount\_trials\_gt\_fixation} = 0.25\,\texttt{events:discount\_amt\_hi\_del\_late} + 0.25\,\texttt{events:discount\_amt\_hi\_del\_soon} + 0.25\,\texttt{events:discount\_amt\_lo\_del\_late} + 0.25\,\texttt{events:discount\_amt\_lo\_del\_soon} - \texttt{events:fixation}. Single conditions: \texttt{events:discount\_amt\_hi\_del\_late}, \texttt{events:discount\_amt\_hi\_del\_soon}, \texttt{events:discount\_amt\_lo\_del\_late}, \texttt{events:discount\_amt\_lo\_del\_soon}.
    \item[DotPatterns.] Contrasts: \texttt{all\_stimuli} = 0.25\,\texttt{events:cue\_a} + 0.25\,\texttt{events:cue\_b} + 0.25\,\texttt{events:probe\_x} + 0.25\,\texttt{events:probe\_y}. Single conditions: \texttt{events:cue\_a}, \texttt{events:cue\_b}, \texttt{events:probe\_x}, \texttt{events:probe\_y}.
    \item[EmoMem.] Contrasts: \texttt{object\_overlay} = \texttt{events:object\_overlay}. Single conditions: \texttt{events:object\_overlay}.
    \item[EmoReco.] Contrasts: \texttt{face\_perception} = 0.25\,\texttt{events:angry\_female\_face} + 0.25\,\texttt{events:angry\_male\_face} + 0.25\,\texttt{events:neutral\_female\_face} + 0.25\,\texttt{events:neutral\_male\_face}. Single conditions: \texttt{events:angry\_female\_face}, \texttt{events:angry\_male\_face}, \texttt{events:neutral\_female\_face}, \texttt{events:neutral\_male\_face}.
    \item[EmotionalPain.] Contrasts: \texttt{pain\_story\_gt\_fixation} = 0.5\,\texttt{events:emotional\_pain\_story} + 0.5\,\texttt{events:physical\_pain\_story} - \texttt{events:fixation}; \texttt{emotional\_pain\_story\_gt\_physical\_pain\_story} = \texttt{events:emotional\_pain\_story} - \texttt{events:physical\_pain\_story}. Single conditions: \texttt{events:emotional\_pain\_story}, \texttt{events:physical\_pain\_story}.
    \item[Enumeration.] Contrasts: \texttt{enumeration\_gt\_fixation} = 0.125\,\texttt{events:enumeration\_response\_1} + 0.125\,\texttt{events:enumeration\_response\_2} + 0.125\,\texttt{events:enumeration\_response\_3} + 0.125\,\texttt{events:enumeration\_response\_4} + 0.125\,\texttt{events:enumeration\_response\_5} + 0.125\,\texttt{events:enumeration\_response\_6} + 0.125\,\texttt{events:enumeration\_response\_7} + 0.125\,\texttt{events:enumeration\_response\_8} - \texttt{events:fixation}; \texttt{high\_gt\_low\_numerosity\_linear} = -0.4375\,\texttt{events:enumeration\_response\_1} - 0.3125\,\texttt{events:enumeration\_response\_2} - 0.1875\,\texttt{events:enumeration\_response\_3} - 0.0625\,\texttt{events:enumeration\_response\_4} + 0.0625\,\texttt{events:enumeration\_response\_5} + 0.1875\,\texttt{events:enumeration\_response\_6} + 0.3125\,\texttt{events:enumeration\_response\_7} + 0.4375\,\texttt{events:enumeration\_response\_8}. Single conditions: \texttt{events:enumeration\_response\_1}, \texttt{events:enumeration\_response\_2}, \texttt{events:enumeration\_response\_3}, \texttt{events:enumeration\_response\_4}, \texttt{events:enumeration\_response\_5}, \texttt{events:enumeration\_response\_6}, \texttt{events:enumeration\_response\_7}, \texttt{events:enumeration\_response\_8}.
    \item[FingerTap.] Contrasts: \texttt{fingertap\_gt\_fixation} = \texttt{events:fingertap} - \texttt{events:fixation}. Single conditions: \texttt{events:fingertap}.
    \item[ItemRecognition.] Contrasts: \texttt{probe\_digit\_decision} = \texttt{events:probe\_digit\_decision}; \texttt{probe\_digit\_decision\_gt\_arrow\_direction\_decision} = \texttt{events:probe\_digit\_decision} - \texttt{events:arrow\_direction\_decision}. Single conditions: \texttt{events:probe\_digit\_decision}, \texttt{events:arrow\_direction\_decision}.
    \item[Lec1.] Contrasts: \texttt{all\_trials\_gt\_fixation} = 0.3333\,\texttt{events:word\_trial} + 0.3333\,\texttt{events:pseudoword\_trial} + 0.3333\,\texttt{events:random\_string\_trial} - \texttt{events:fixation}; \texttt{word\_gt\_pseudoword\_and\_random\_string} = \texttt{events:word\_trial} - 0.5\,\texttt{events:pseudoword\_trial} - 0.5\,\texttt{events:random\_string\_trial}. Single conditions: \texttt{events:word\_trial}, \texttt{events:pseudoword\_trial}, \texttt{events:random\_string\_trial}.
    \item[MCSE.] Contrasts: \texttt{search\_gt\_fixation} = 0.25\,\texttt{events:high\_salience\_left} + 0.25\,\texttt{events:high\_salience\_right} + 0.25\,\texttt{events:low\_salience\_left} + 0.25\,\texttt{events:low\_salience\_right} - \texttt{events:fixation}; \texttt{low\_salience\_gt\_high\_salience} = 0.5\,\texttt{events:low\_salience\_left} + 0.5\,\texttt{events:low\_salience\_right} - 0.5\,\texttt{events:high\_salience\_left} - 0.5\,\texttt{events:high\_salience\_right}. Single conditions: \texttt{events:high\_salience\_left}, \texttt{events:high\_salience\_right}, \texttt{events:low\_salience\_left}, \texttt{events:low\_salience\_right}.
    \item[MTTNS.] Contrasts: \texttt{all\_events\_gt\_fixation} = 0.125\,\texttt{events:after\_close\_event} + 0.125\,\texttt{events:after\_far\_event} + 0.125\,\texttt{events:before\_close\_event} + 0.125\,\texttt{events:before\_far\_event} + 0.125\,\texttt{events:northside\_close\_event} + 0.125\,\texttt{events:northside\_far\_event} + 0.125\,\texttt{events:southside\_close\_event} + 0.125\,\texttt{events:southside\_far\_event} - \texttt{events:fixation}; \texttt{space\_event\_gt\_time\_event} = 0.25\,\texttt{events:northside\_close\_event} + 0.25\,\texttt{events:northside\_far\_event} + 0.25\,\texttt{events:southside\_close\_event} + 0.25\,\texttt{events:southside\_far\_event} - 0.25\,\texttt{events:after\_close\_event} - 0.25\,\texttt{events:after\_far\_event} - 0.25\,\texttt{events:before\_close\_event} - 0.25\,\texttt{events:before\_far\_event}. Single conditions: \texttt{events:after\_close\_event}, \texttt{events:after\_far\_event}, \texttt{events:before\_close\_event}, \texttt{events:before\_far\_event}, \texttt{events:northside\_close\_event}, \texttt{events:northside\_far\_event}, \texttt{events:southside\_close\_event}, \texttt{events:southside\_far\_event}.
    \item[MTTWE.] Contrasts: \texttt{all\_event\_audio\_gt\_fixation} = 0.25\,\texttt{events:after\_event\_audio} + 0.25\,\texttt{events:before\_event\_audio} + 0.25\,\texttt{events:eastside\_event\_audio} + 0.25\,\texttt{events:westside\_event\_audio} - \texttt{events:fixation}; \texttt{time\_event\_audio\_gt\_space\_event\_audio} = 0.5\,\texttt{events:after\_event\_audio} + 0.5\,\texttt{events:before\_event\_audio} - 0.5\,\texttt{events:eastside\_event\_audio} - 0.5\,\texttt{events:westside\_event\_audio}. Single conditions: \texttt{events:after\_event\_audio}, \texttt{events:before\_event\_audio}, \texttt{events:eastside\_event\_audio}, \texttt{events:westside\_event\_audio}.
    \item[MVEB.] Contrasts: \texttt{string\_encoding\_gt\_fixation} = 0.1667\,\texttt{events:two\_letters\_different\_string} + 0.1667\,\texttt{events:two\_letters\_same\_string} + 0.1667\,\texttt{events:four\_letters\_different\_string} + 0.1667\,\texttt{events:four\_letters\_same\_string} + 0.1667\,\texttt{events:six\_letters\_different\_string} + 0.1667\,\texttt{events:six\_letters\_same\_string} - \texttt{events:fixation}. Single conditions: \texttt{events:two\_letters\_different\_string}, \texttt{events:two\_letters\_same\_string}, \texttt{events:four\_letters\_different\_string}, \texttt{events:four\_letters\_same\_string}, \texttt{events:six\_letters\_different\_string}, \texttt{events:six\_letters\_same\_string}.
    \item[MVIS.] Contrasts: \texttt{task\_gt\_fixation} = 0.125\,\texttt{events:empty\_grid} + 0.125\,\texttt{events:probe\_dot\_decision} + 0.125\,\texttt{events:two\_dots\_array} + 0.125\,\texttt{events:two\_dots\_control\_array} + 0.125\,\texttt{events:four\_dots\_array} + 0.125\,\texttt{events:four\_dots\_control\_array} + 0.125\,\texttt{events:six\_dots\_array} + 0.125\,\texttt{events:six\_dots\_control\_array} - \texttt{events:fixation}. Single conditions: \texttt{events:empty\_grid}, \texttt{events:probe\_dot\_decision}, \texttt{events:two\_dots\_array}, \texttt{events:two\_dots\_control\_array}, \texttt{events:four\_dots\_array}, \texttt{events:four\_dots\_control\_array}, \texttt{events:six\_dots\_array}, \texttt{events:six\_dots\_control\_array}.
    \item[MathLanguage.] Contrasts: \texttt{task\_stimuli} = 0.08333\,\texttt{events:arithmetic\_fact\_auditory} + 0.08333\,\texttt{events:arithmetic\_fact\_visual} + 0.08333\,\texttt{events:arithmetic\_principle\_auditory} + 0.08333\,\texttt{events:arithmetic\_principle\_visual} + 0.08333\,\texttt{events:colorlessg\_auditory} + 0.08333\,\texttt{events:colorlessg\_visual} + 0.08333\,\texttt{events:general\_auditory} + 0.08333\,\texttt{events:general\_visual} + 0.08333\,\texttt{events:geometry\_fact\_auditory} + 0.08333\,\texttt{events:geometry\_fact\_visual} + 0.08333\,\texttt{events:theory\_of\_mind\_auditory} + 0.08333\,\texttt{events:theory\_of\_mind\_visual}. Single conditions: \texttt{events:arithmetic\_fact\_auditory}, \texttt{events:arithmetic\_fact\_visual}, \texttt{events:arithmetic\_principle\_auditory}, \texttt{events:arithmetic\_principle\_visual}, \texttt{events:colorlessg\_auditory}, \texttt{events:colorlessg\_visual}, \texttt{events:general\_auditory}, \texttt{events:general\_visual}, \texttt{events:geometry\_fact\_auditory}, \texttt{events:geometry\_fact\_visual}, \texttt{events:theory\_of\_mind\_auditory}, \texttt{events:theory\_of\_mind\_visual}.
    \item[Moto.] Contrasts: \texttt{saccade\_gt\_other\_movements} = 0.5\,\texttt{events:saccade\_left} + 0.5\,\texttt{events:saccade\_right} - 0.125\,\texttt{events:finger\_left} - 0.125\,\texttt{events:finger\_right} - 0.125\,\texttt{events:foot\_left} - 0.125\,\texttt{events:foot\_right} - 0.125\,\texttt{events:hand\_left} - 0.125\,\texttt{events:hand\_right} - 0.125\,\texttt{events:tongue\_left} - 0.125\,\texttt{events:tongue\_right}. Single conditions: \texttt{events:saccade\_left}, \texttt{events:saccade\_right}, \texttt{events:finger\_left}, \texttt{events:finger\_right}, \texttt{events:foot\_left}, \texttt{events:foot\_right}, \texttt{events:hand\_left}, \texttt{events:hand\_right}, \texttt{events:tongue\_left}, \texttt{events:tongue\_right}.
    \item[NARPS.] Contrasts: \texttt{stim\_gt\_fixation} = 0.25\,\texttt{events:narps\_gamble\_gain\_hi\_loss\_hi} + 0.25\,\texttt{events:narps\_gamble\_gain\_hi\_loss\_lo} + 0.25\,\texttt{events:narps\_gamble\_gain\_lo\_loss\_hi} + 0.25\,\texttt{events:narps\_gamble\_gain\_lo\_loss\_lo} - \texttt{events:fixation}. Single conditions: \texttt{events:narps\_gamble\_gain\_hi\_loss\_hi}, \texttt{events:narps\_gamble\_gain\_hi\_loss\_lo}, \texttt{events:narps\_gamble\_gain\_lo\_loss\_hi}, \texttt{events:narps\_gamble\_gain\_lo\_loss\_lo}.
    \item[PainMovie.] Contrasts: \texttt{movie\_scenes\_gt\_fixation} = 0.5\,\texttt{events:mental\_movie\_scene} + 0.5\,\texttt{events:pain\_movie\_scene} - \texttt{events:fixation}. Single conditions: \texttt{events:mental\_movie\_scene}, \texttt{events:pain\_movie\_scene}.
    \item[PreferenceFaces.] Contrasts: \texttt{face\_evaluation\_gt\_fixation} = 0.5\,\texttt{events:face\_score\_hi} + 0.5\,\texttt{events:face\_score\_lo} - \texttt{events:fixation}. Single conditions: \texttt{events:face\_score\_hi}, \texttt{events:face\_score\_lo}.
    \item[PreferenceFood.] Contrasts: \texttt{food\_gt\_fixation} = 0.5\,\texttt{events:food\_score\_hi} + 0.5\,\texttt{events:food\_score\_lo} - \texttt{events:fixation}. Single conditions: \texttt{events:food\_score\_hi}, \texttt{events:food\_score\_lo}.
    \item[PreferenceHouses.] Contrasts: \texttt{houses\_gt\_fixation} = 0.5\,\texttt{events:house\_score\_hi} + 0.5\,\texttt{events:house\_score\_lo} - \texttt{events:fixation}. Single conditions: \texttt{events:house\_score\_hi}, \texttt{events:house\_score\_lo}.
    \item[PreferencePaintings.] Contrasts: \texttt{paintings\_gt\_fixation} = 0.5\,\texttt{events:painting\_score\_hi} + 0.5\,\texttt{events:painting\_score\_lo} - \texttt{events:fixation}. Single conditions: \texttt{events:painting\_score\_hi}, \texttt{events:painting\_score\_lo}.
    \item[RSVPLanguage.] Contrasts: \texttt{reading\_blocks\_gt\_fixation} = 0.1667\,\texttt{events:complex\_sentence} + 0.1667\,\texttt{events:simple\_sentence} + 0.1667\,\texttt{events:word\_list} + 0.1667\,\texttt{events:jabberwocky} + 0.1667\,\texttt{events:pseudoword\_list} + 0.1667\,\texttt{events:consonant\_strings} - \texttt{events:fixation}. Single conditions: \texttt{events:complex\_sentence}, \texttt{events:simple\_sentence}, \texttt{events:word\_list}, \texttt{events:jabberwocky}, \texttt{events:pseudoword\_list}, \texttt{events:consonant\_strings}.
    \item[SelectiveStopSignal.] Contrasts: \texttt{task\_gt\_fixation} = 0.25\,\texttt{events:critical\_go} + 0.25\,\texttt{events:critical\_stop} + 0.25\,\texttt{events:noncritical\_go} + 0.25\,\texttt{events:noncritical\_ignore} - \texttt{events:fixation}; \texttt{critical\_stop\_gt\_noncritical\_ignore} = \texttt{events:critical\_stop} - \texttt{events:noncritical\_ignore}. Single conditions: \texttt{events:critical\_go}, \texttt{events:critical\_stop}, \texttt{events:noncritical\_go}, \texttt{events:noncritical\_ignore}.
    \item[Self.] Contrasts: \texttt{task\_gt\_fixation} = 0.25\,\texttt{events:other\_adjective\_judgment} + 0.25\,\texttt{events:self\_adjective\_judgment} + 0.25\,\texttt{events:recognition\_yes} + 0.25\,\texttt{events:recognition\_no} - \texttt{events:fixation}. Single conditions: \texttt{events:other\_adjective\_judgment}, \texttt{events:self\_adjective\_judgment}, \texttt{events:recognition\_yes}, \texttt{events:recognition\_no}.
    \item[SpatialNavigation.] Contrasts: \texttt{encoding\_phase} = 0.5\,\texttt{events:encoding\_intersection} + 0.5\,\texttt{events:encoding\_navigation}; \texttt{experimental\_gt\_control} = \texttt{events:experimental\_approach} - \texttt{events:control\_approach}. Single conditions: \texttt{events:encoding\_intersection}, \texttt{events:encoding\_navigation}, \texttt{events:experimental\_approach}, \texttt{events:control\_approach}.
    \item[StopSignal.] Contrasts: \texttt{task\_trials} = 0.5\,\texttt{events:go\_trial} + 0.5\,\texttt{events:stop\_trial}; \texttt{stop\_trial\_gt\_go\_trial} = \texttt{events:stop\_trial} - \texttt{events:go\_trial}. Single conditions: \texttt{events:go\_trial}, \texttt{events:stop\_trial}.
    \item[Stroop.] Contrasts: \texttt{task\_all} = 0.5\,\texttt{events:congruent} + 0.5\,\texttt{events:incongruent}. Single conditions: \texttt{events:congruent}, \texttt{events:incongruent}.
    \item[TheoryOfMind.] Contrasts: \texttt{task\_gt\_fixation} = 0.25\,\texttt{events:belief\_story} + 0.25\,\texttt{events:belief\_question} + 0.25\,\texttt{events:photo\_story} + 0.25\,\texttt{events:photo\_question} - \texttt{events:fixation}; \texttt{belief\_gt\_photo} = 0.5\,\texttt{events:belief\_story} + 0.5\,\texttt{events:belief\_question} - 0.5\,\texttt{events:photo\_story} - 0.5\,\texttt{events:photo\_question}. Single conditions: \texttt{events:belief\_story}, \texttt{events:belief\_question}, \texttt{events:photo\_story}, \texttt{events:photo\_question}.
    \item[TwoByTwo.] Contrasts: \texttt{task\_stimulus\_gt\_fixation} = \texttt{events:task\_stimulus} - \texttt{events:fixation}. Single conditions: \texttt{events:task\_stimulus}.
    \item[VSTM.] Contrasts: \texttt{all\_stimuli\_gt\_fixation} = 0.08333\,\texttt{events:vstm\_encode\_load\_1} + 0.08333\,\texttt{events:vstm\_encode\_load\_2} + 0.08333\,\texttt{events:vstm\_encode\_load\_3} + 0.08333\,\texttt{events:vstm\_encode\_load\_4} + 0.08333\,\texttt{events:vstm\_encode\_load\_5} + 0.08333\,\texttt{events:vstm\_encode\_load\_6} + 0.08333\,\texttt{events:vstm\_probe\_load\_1} + 0.08333\,\texttt{events:vstm\_probe\_load\_2} + 0.08333\,\texttt{events:vstm\_probe\_load\_3} + 0.08333\,\texttt{events:vstm\_probe\_load\_4} + 0.08333\,\texttt{events:vstm\_probe\_load\_5} + 0.08333\,\texttt{events:vstm\_probe\_load\_6} - \texttt{events:fixation}. Single conditions: \texttt{events:vstm\_encode\_load\_1}, \texttt{events:vstm\_encode\_load\_2}, \texttt{events:vstm\_encode\_load\_3}, \texttt{events:vstm\_encode\_load\_4}, \texttt{events:vstm\_encode\_load\_5}, \texttt{events:vstm\_encode\_load\_6}, \texttt{events:vstm\_probe\_load\_1}, \texttt{events:vstm\_probe\_load\_2}, \texttt{events:vstm\_probe\_load\_3}, \texttt{events:vstm\_probe\_load\_4}, \texttt{events:vstm\_probe\_load\_5}, \texttt{events:vstm\_probe\_load\_6}.
    \item[VSTMC.] Contrasts: \texttt{stim} = 0.3333\,\texttt{events:stimulus\_load1} + 0.3333\,\texttt{events:stimulus\_load2} + 0.3333\,\texttt{events:stimulus\_load3}. Single conditions: \texttt{events:stimulus\_load1}, \texttt{events:stimulus\_load2}, \texttt{events:stimulus\_load3}.
    \item[VisualSearch.] Contrasts: \texttt{task\_stimuli} = 0.09091\,\texttt{events:sample\_item} + 0.09091\,\texttt{events:memory\_array\_two} + 0.09091\,\texttt{events:memory\_array\_four} + 0.09091\,\texttt{events:probe\_item\_two\_absent} + 0.09091\,\texttt{events:probe\_item\_two\_present} + 0.09091\,\texttt{events:probe\_item\_four\_absent} + 0.09091\,\texttt{events:probe\_item\_four\_present} + 0.09091\,\texttt{events:search\_array\_two\_absent} + 0.09091\,\texttt{events:search\_array\_two\_present} + 0.09091\,\texttt{events:search\_array\_four\_absent} + 0.09091\,\texttt{events:search\_array\_four\_present}. Single conditions: \texttt{events:sample\_item}, \texttt{events:memory\_array\_two}, \texttt{events:memory\_array\_four}, \texttt{events:probe\_item\_two\_absent}, \texttt{events:probe\_item\_two\_present}, \texttt{events:probe\_item\_four\_absent}, \texttt{events:probe\_item\_four\_present}, \texttt{events:search\_array\_two\_absent}, \texttt{events:search\_array\_two\_present}, \texttt{events:search\_array\_four\_absent}, \texttt{events:search\_array\_four\_present}.
    \item[WardAndAllport.] Contrasts: \texttt{planning\_problem\_gt\_fixation} = \texttt{events:planning\_problem} - \texttt{events:fixation}; \texttt{planning\_problem\_gt\_move\_execution} = \texttt{events:planning\_problem} - \texttt{events:move\_execution}. Single conditions: \texttt{events:planning\_problem}, \texttt{events:move\_execution}.
\end{description}
}

IBC total: 60 contrasts and 191 single conditions.

\section{LLM Use}
\label{app:llm}

We use LLMs to generate and rephrase captions of task instructions, sensory presentations, and responses. A mixture of GPT-5.4-codex by OpenAI and Claude Opus 4.6 by Anthropic were used for this purpose.

\newpage
\section*{NeurIPS Paper Checklist}

\begin{enumerate}

\item {\bf Claims}
    \item[] Question: Do the main claims made in the abstract and introduction accurately reflect the paper's contributions and scope?
    \item[] Answer: \answerYes{} 
    \item[] Justification: The abstract and introduction state the model, conditioning regimes, zero-shot scope, and intended use cases. The Methods and Results sections evaluate these claims on held-out tasks across contrast recovery, distance-to-support analyses, time-series statistics, and counterfactual timing experiments.
    \item[] Guidelines:
    \begin{itemize}
        \item The answer \answerNA{} means that the abstract and introduction do not include the claims made in the paper.
        \item The abstract and/or introduction should clearly state the claims made, including the contributions made in the paper and important assumptions and limitations. A \answerNo{} or \answerNA{} answer to this question will not be perceived well by the reviewers. 
        \item The claims made should match theoretical and experimental results, and reflect how much the results can be expected to generalize to other settings. 
        \item It is fine to include aspirational goals as motivation as long as it is clear that these goals are not attained by the paper. 
    \end{itemize}

\item {\bf Limitations}
    \item[] Question: Does the paper discuss the limitations of the work performed by the authors?
    \item[] Answer: \answerYes{} 
    \item[] Justification: The paper includes a Limitations section discussing the scope of extrapolation, assumptions behind compositional generalization, oracle-like spatial priors, and limits of language-only conditioning.
    \item[] Guidelines:
    \begin{itemize}
        \item The answer \answerNA{} means that the paper has no limitation while the answer \answerNo{} means that the paper has limitations, but those are not discussed in the paper. 
        \item The authors are encouraged to create a separate ``Limitations'' section in their paper.
        \item The paper should point out any strong assumptions and how robust the results are to violations of these assumptions (e.g., independence assumptions, noiseless settings, model well-specification, asymptotic approximations only holding locally). The authors should reflect on how these assumptions might be violated in practice and what the implications would be.
        \item The authors should reflect on the scope of the claims made, e.g., if the approach was only tested on a few datasets or with a few runs. In general, empirical results often depend on implicit assumptions, which should be articulated.
        \item The authors should reflect on the factors that influence the performance of the approach. For example, a facial recognition algorithm may perform poorly when image resolution is low or images are taken in low lighting. Or a speech-to-text system might not be used reliably to provide closed captions for online lectures because it fails to handle technical jargon.
        \item The authors should discuss the computational efficiency of the proposed algorithms and how they scale with dataset size.
        \item If applicable, the authors should discuss possible limitations of their approach to address problems of privacy and fairness.
        \item While the authors might fear that complete honesty about limitations might be used by reviewers as grounds for rejection, a worse outcome might be that reviewers discover limitations that aren't acknowledged in the paper. The authors should use their best judgment and recognize that individual actions in favor of transparency play an important role in developing norms that preserve the integrity of the community. Reviewers will be specifically instructed to not penalize honesty concerning limitations.
    \end{itemize}

\item {\bf Theory assumptions and proofs}
    \item[] Question: For each theoretical result, does the paper provide the full set of assumptions and a complete (and correct) proof?
    \item[] Answer: \answerNA{} 
    \item[] Justification: The paper uses mathematical notation to define the model and flow-matching objective, but it does not introduce new theoretical results, theorems, or proofs
    \item[] Guidelines:
    \begin{itemize}
        \item The answer \answerNA{} means that the paper does not include theoretical results. 
        \item All the theorems, formulas, and proofs in the paper should be numbered and cross-referenced.
        \item All assumptions should be clearly stated or referenced in the statement of any theorems.
        \item The proofs can either appear in the main paper or the supplemental material, but if they appear in the supplemental material, the authors are encouraged to provide a short proof sketch to provide intuition. 
        \item Inversely, any informal proof provided in the core of the paper should be complemented by formal proofs provided in appendix or supplemental material.
        \item Theorems and Lemmas that the proof relies upon should be properly referenced. 
    \end{itemize}

    \item {\bf Experimental result reproducibility}
    \item[] Question: Does the paper fully disclose all the information needed to reproduce the main experimental results of the paper to the extent that it affects the main claims and/or conclusions of the paper (regardless of whether the code and data are provided or not)?
    \item[] Answer: \answerYes{} 
    \item[] Justification: The Methods and Experimental Setup sections describe the architecture, conditioning tokens, flow-matching objective, datasets, preprocessing, held-out folds, GLM evaluation, baselines, hyperparameters, and compute budget; additional optimization, dropout, sampling, and evaluation details are referenced in the appendix. There we also describe how to access the data used.

    \item[] Guidelines:
    \begin{itemize}
        \item The answer \answerNA{} means that the paper does not include experiments.
        \item If the paper includes experiments, a \answerNo{} answer to this question will not be perceived well by the reviewers: Making the paper reproducible is important, regardless of whether the code and data are provided or not.
        \item If the contribution is a dataset and\slash or model, the authors should describe the steps taken to make their results reproducible or verifiable. 
        \item Depending on the contribution, reproducibility can be accomplished in various ways. For example, if the contribution is a novel architecture, describing the architecture fully might suffice, or if the contribution is a specific model and empirical evaluation, it may be necessary to either make it possible for others to replicate the model with the same dataset, or provide access to the model. In general. releasing code and data is often one good way to accomplish this, but reproducibility can also be provided via detailed instructions for how to replicate the results, access to a hosted model (e.g., in the case of a large language model), releasing of a model checkpoint, or other means that are appropriate to the research performed.
        \item While NeurIPS does not require releasing code, the conference does require all submissions to provide some reasonable avenue for reproducibility, which may depend on the nature of the contribution. For example
        \begin{enumerate}
            \item If the contribution is primarily a new algorithm, the paper should make it clear how to reproduce that algorithm.
            \item If the contribution is primarily a new model architecture, the paper should describe the architecture clearly and fully.
            \item If the contribution is a new model (e.g., a large language model), then there should either be a way to access this model for reproducing the results or a way to reproduce the model (e.g., with an open-source dataset or instructions for how to construct the dataset).
            \item We recognize that reproducibility may be tricky in some cases, in which case authors are welcome to describe the particular way they provide for reproducibility. In the case of closed-source models, it may be that access to the model is limited in some way (e.g., to registered users), but it should be possible for other researchers to have some path to reproducing or verifying the results.
        \end{enumerate}
    \end{itemize}

\item {\bf Open access to data and code}
    \item[] Question: Does the paper provide open access to the data and code, with sufficient instructions to faithfully reproduce the main experimental results, as described in supplemental material?
    \item[] Answer: \answerNo{} 
    \item[] Justification: The paper uses controlled-access neuroimaging datasets whose raw data must be requested from the original providers under dataset-specific data-use agreements. We provide data-access information and implementation details in the supplement, but do not include a fully reproducible code and data release at submission time. We will release the training/evaluation code, preprocessing scripts, and model weights upon acceptance, subject to the redistribution terms of the original datasets.
    \item[] Guidelines:
    \begin{itemize}
        \item The answer \answerNA{} means that paper does not include experiments requiring code.
        \item Please see the NeurIPS code and data submission guidelines (\url{https://neurips.cc/public/guides/CodeSubmissionPolicy}) for more details.
        \item While we encourage the release of code and data, we understand that this might not be possible, so \answerNo{} is an acceptable answer. Papers cannot be rejected simply for not including code, unless this is central to the contribution (e.g., for a new open-source benchmark).
        \item The instructions should contain the exact command and environment needed to run to reproduce the results. See the NeurIPS code and data submission guidelines (\url{https://neurips.cc/public/guides/CodeSubmissionPolicy}) for more details.
        \item The authors should provide instructions on data access and preparation, including how to access the raw data, preprocessed data, intermediate data, and generated data, etc.
        \item The authors should provide scripts to reproduce all experimental results for the new proposed method and baselines. If only a subset of experiments are reproducible, they should state which ones are omitted from the script and why.
        \item At submission time, to preserve anonymity, the authors should release anonymized versions (if applicable).
        \item Providing as much information as possible in supplemental material (appended to the paper) is recommended, but including URLs to data and code is permitted.
    \end{itemize}

\item {\bf Experimental setting/details}
    \item[] Question: Does the paper specify all the training and test details (e.g., data splits, hyperparameters, how they were chosen, type of optimizer) necessary to understand the results?
    \item[] Answer: \answerYes{} 
    \item[] Justification: The main paper specifies the datasets, held-out-task and held-out-subject
  splits, model architecture, conditioning pathways, evaluation protocol, main hyperparameters, and
  baselines needed to interpret the results. Additional optimization details,
sampling weights, dropout probabilities, contrast definitions, and dataset-specific evaluation
  details are provided in the appendix.
    \item[] Guidelines:
    \begin{itemize}
        \item The answer \answerNA{} means that the paper does not include experiments.
        \item The experimental setting should be presented in the core of the paper to a level of detail that is necessary to appreciate the results and make sense of them.
        \item The full details can be provided either with the code, in appendix, or as supplemental material.
    \end{itemize}

\item {\bf Experiment statistical significance}
    \item[] Question: Does the paper report error bars suitably and correctly defined or other appropriate information about the statistical significance of the experiments?
    \item[] Answer: \answerYes{} 
    \item[] Justification: For all instances where errors are reported, we clarify that they are standard deviations. We use statistical testing to support two of our claims and indicate the test and over which data they were computed.
    \item[] Guidelines:
    \begin{itemize}
        \item The answer \answerNA{} means that the paper does not include experiments.
        \item The authors should answer \answerYes{} if the results are accompanied by error bars, confidence intervals, or statistical significance tests, at least for the experiments that support the main claims of the paper.
        \item The factors of variability that the error bars are capturing should be clearly stated (for example, train/test split, initialization, random drawing of some parameter, or overall run with given experimental conditions).
        \item The method for calculating the error bars should be explained (closed form formula, call to a library function, bootstrap, etc.)
        \item The assumptions made should be given (e.g., Normally distributed errors).
        \item It should be clear whether the error bar is the standard deviation or the standard error of the mean.
        \item It is OK to report 1-sigma error bars, but one should state it. The authors should preferably report a 2-sigma error bar than state that they have a 96\% CI, if the hypothesis of Normality of errors is not verified.
        \item For asymmetric distributions, the authors should be careful not to show in tables or figures symmetric error bars that would yield results that are out of range (e.g., negative error rates).
        \item If error bars are reported in tables or plots, the authors should explain in the text how they were calculated and reference the corresponding figures or tables in the text.
    \end{itemize}

\item {\bf Experiments compute resources}
    \item[] Question: For each experiment, does the paper provide sufficient information on the computer resources (type of compute workers, memory, time of execution) needed to reproduce the experiments?
    \item[] Answer: \answerYes{} 
    \item[] Justification: We include the compute use in its own Appendix section and explicitly refer to this in the main text.
    \item[] Guidelines:
    \begin{itemize}
        \item The answer \answerNA{} means that the paper does not include experiments.
        \item The paper should indicate the type of compute workers CPU or GPU, internal cluster, or cloud provider, including relevant memory and storage.
        \item The paper should provide the amount of compute required for each of the individual experimental runs as well as estimate the total compute. 
        \item The paper should disclose whether the full research project required more compute than the experiments reported in the paper (e.g., preliminary or failed experiments that didn't make it into the paper). 
    \end{itemize}
    
\item {\bf Code of ethics}
    \item[] Question: Does the research conducted in the paper conform, in every respect, with the NeurIPS Code of Ethics \url{https://neurips.cc/public/EthicsGuidelines}?
    \item[] Answer: \answerYes{} 
    \item[] Justification: We confirm to have read and conducted our research fully in line with the NeurIPS Code of Ethics.
    \item[] Guidelines:
    \begin{itemize}
        \item The answer \answerNA{} means that the authors have not reviewed the NeurIPS Code of Ethics.
        \item If the authors answer \answerNo, they should explain the special circumstances that require a deviation from the Code of Ethics.
        \item The authors should make sure to preserve anonymity (e.g., if there is a special consideration due to laws or regulations in their jurisdiction).
    \end{itemize}

\item {\bf Broader impacts}
    \item[] Question: Does the paper discuss both potential positive societal impacts and negative societal impacts of the work performed?
    \item[] Answer: \answerYes{} 
    \item[] Justification: The paper discusses positive impacts for in-silico task design and counterfactual neuroscience, as well as limitations and risks related to over-interpreting generated brain responses, subject-identifiable neuroimaging structure, and the need for controlled use of individual-level data.
    \item[] Guidelines:
    \begin{itemize}
        \item The answer \answerNA{} means that there is no societal impact of the work performed.
        \item If the authors answer \answerNA{} or \answerNo, they should explain why their work has no societal impact or why the paper does not address societal impact.
        \item Examples of negative societal impacts include potential malicious or unintended uses (e.g., disinformation, generating fake profiles, surveillance), fairness considerations (e.g., deployment of technologies that could make decisions that unfairly impact specific groups), privacy considerations, and security considerations.
        \item The conference expects that many papers will be foundational research and not tied to particular applications, let alone deployments. However, if there is a direct path to any negative applications, the authors should point it out. For example, it is legitimate to point out that an improvement in the quality of generative models could be used to generate Deepfakes for disinformation. On the other hand, it is not needed to point out that a generic algorithm for optimizing neural networks could enable people to train models that generate Deepfakes faster.
        \item The authors should consider possible harms that could arise when the technology is being used as intended and functioning correctly, harms that could arise when the technology is being used as intended but gives incorrect results, and harms following from (intentional or unintentional) misuse of the technology.
        \item If there are negative societal impacts, the authors could also discuss possible mitigation strategies (e.g., gated release of models, providing defenses in addition to attacks, mechanisms for monitoring misuse, mechanisms to monitor how a system learns from feedback over time, improving the efficiency and accessibility of ML).
    \end{itemize}
    
\item {\bf Safeguards}
    \item[] Question: Does the paper describe safeguards that have been put in place for responsible release of data or models that have a high risk for misuse (e.g., pre-trained language models, image generators, or scraped datasets)?
    \item[] Answer: \answerNA{} 
    \item[] Justification: The paper does not release a high-risk model or dataset at submission time. Raw neuroimaging data remain governed by the original providers' controlled-access agreements; any
  future release of code, weights, or generated data will follow these terms.
    \item[] Guidelines:
    \begin{itemize}
        \item The answer \answerNA{} means that the paper poses no such risks.
        \item Released models that have a high risk for misuse or dual-use should be released with necessary safeguards to allow for controlled use of the model, for example by requiring that users adhere to usage guidelines or restrictions to access the model or implementing safety filters. 
        \item Datasets that have been scraped from the Internet could pose safety risks. The authors should describe how they avoided releasing unsafe images.
        \item We recognize that providing effective safeguards is challenging, and many papers do not require this, but we encourage authors to take this into account and make a best faith effort.
    \end{itemize}

\item {\bf Licenses for existing assets}
    \item[] Question: Are the creators or original owners of assets (e.g., code, data, models), used in the paper, properly credited and are the license and terms of use explicitly mentioned and properly respected?
    \item[] Answer: \answerYes{} 
    \item[] Justification: The paper cites all existing datasets and model components used. Dataset access is governed by the respective providers' data-use agreements; raw data are not redistributed. Code and model dependencies will be documented in the release package.
    \item[] Guidelines:
    \begin{itemize}
        \item The answer \answerNA{} means that the paper does not use existing assets.
        \item The authors should cite the original paper that produced the code package or dataset.
        \item The authors should state which version of the asset is used and, if possible, include a URL.
        \item The name of the license (e.g., CC-BY 4.0) should be included for each asset.
        \item For scraped data from a particular source (e.g., website), the copyright and terms of service of that source should be provided.
        \item If assets are released, the license, copyright information, and terms of use in the package should be provided. For popular datasets, \url{paperswithcode.com/datasets} has curated licenses for some datasets. Their licensing guide can help determine the license of a dataset.
        \item For existing datasets that are re-packaged, both the original license and the license of the derived asset (if it has changed) should be provided.
        \item If this information is not available online, the authors are encouraged to reach out to the asset's creators.
    \end{itemize}

\item {\bf New assets}
    \item[] Question: Are new assets introduced in the paper well documented and is the documentation provided alongside the assets?
    \item[] Answer: \answerNo{} 
    \item[] Justification: No new public dataset, code package, or model weights are released with the initial submission. Upon acceptance, we plan to release code, configuration files, preprocessing scripts, and trained model weights, subject to the redistribution terms of the source datasets.
    \item[] Guidelines:
    \begin{itemize}
        \item The answer \answerNA{} means that the paper does not release new assets.
        \item Researchers should communicate the details of the dataset\slash code\slash model as part of their submissions via structured templates. This includes details about training, license, limitations, etc. 
        \item The paper should discuss whether and how consent was obtained from people whose asset is used.
        \item At submission time, remember to anonymize your assets (if applicable). You can either create an anonymized URL or include an anonymized zip file.
    \end{itemize}

\item {\bf Crowdsourcing and research with human subjects}
    \item[] Question: For crowdsourcing experiments and research with human subjects, does the paper include the full text of instructions given to participants and screenshots, if applicable, as well as details about compensation (if any)? 
    \item[] Answer: \answerNA{} 
    \item[] Justification: The paper does not conduct new crowdsourcing experiments or collect new human-subject data. It analyzes previously collected neuroimaging datasets
    \item[] Guidelines:
    \begin{itemize}
        \item The answer \answerNA{} means that the paper does not involve crowdsourcing nor research with human subjects.
        \item Including this information in the supplemental material is fine, but if the main contribution of the paper involves human subjects, then as much detail as possible should be included in the main paper. 
        \item According to the NeurIPS Code of Ethics, workers involved in data collection, curation, or other labor should be paid at least the minimum wage in the country of the data collector. 
    \end{itemize}

\item {\bf Institutional review board (IRB) approvals or equivalent for research with human subjects}
    \item[] Question: Does the paper describe potential risks incurred by study participants, whether such risks were disclosed to the subjects, and whether Institutional Review Board (IRB) approvals (or an equivalent approval/review based on the requirements of your country or institution) were obtained?
    \item[] Answer: \answerNA{} 
    \item[] Justification: The paper does not collect new human-subject data. All analyses use previously collected datasets accessed under their respective data-use agreements and original ethical approvals.
    \item[] Guidelines:
    \begin{itemize}
        \item The answer \answerNA{} means that the paper does not involve crowdsourcing nor research with human subjects.
        \item Depending on the country in which research is conducted, IRB approval (or equivalent) may be required for any human subjects research. If you obtained IRB approval, you should clearly state this in the paper. 
        \item We recognize that the procedures for this may vary significantly between institutions and locations, and we expect authors to adhere to the NeurIPS Code of Ethics and the guidelines for their institution. 
        \item For initial submissions, do not include any information that would break anonymity (if applicable), such as the institution conducting the review.
    \end{itemize}

\item {\bf Declaration of LLM usage}
    \item[] Question: Does the paper describe the usage of LLMs if it is an important, original, or non-standard component of the core methods in this research? Note that if the LLM is used only for writing, editing, or formatting purposes and does \emph{not} impact the core methodology, scientific rigor, or originality of the research, declaration is not required.
    \item[] Answer: \answerYes{} 
    \item[] Justification: The method uses an LLM to generate phrase variants of task captions for conditioning robustness. The paper describes this use in the appendix. LLMs were not used to generate experimental results beyond these caption variants.
    \item[] Guidelines:
    \begin{itemize}
        \item The answer \answerNA{} means that the core method development in this research does not involve LLMs as any important, original, or non-standard components.
        \item Please refer to our LLM policy in the NeurIPS handbook for what should or should not be described.
    \end{itemize}

\end{enumerate}

\end{document}